\documentclass[10pt,twocolumn,letterpaper]{article}

\usepackage{cvpr}      

\definecolor{cvprblue}{rgb}{0.21,0.49,0.74}
\usepackage[pagebackref,breaklinks,colorlinks,allcolors=cvprblue]{hyperref}

\usepackage{algorithm}
\usepackage{algorithmic}
\usepackage{enumitem}
\usepackage{amsmath}
\usepackage{multirow}
\usepackage{booktabs}
\usepackage{pifont}
\usepackage{colortbl}
\usepackage{makecell}

\usepackage{xr}

\newcommand{\xmark}{\ding{55}}\definecolor{lightgray}{gray}{0.9}

\def\paperID{17805} \def\confName{CVPR}
\def\confYear{2026}
\definecolor{delta}{RGB}{0, 128, 0}

\title{Understanding Temporal Logic Consistency in Video-Language Models through Cross-Modal Attention Discriminability}

\author{
    Chengzhi Li$^{1}$ \quad Heyan Huang$^{1,2}$ \quad Ping Jian$^{1,2, \dagger}$ \quad Zhen Yang$^{1}$ \quad Yaning Tian$^{1}$ \quad Zhongbin Guo$^{1}$\\
    $^{1}$School of Computer Science and Technology, Beijing Institute of Technology, Beijing, China\\
    $^{2}$Beijing Engineering Research Center of High Volume Language Information Processing \\ and Cloud Computing Applications, Beijing Institute of Technology, Beijing, China \\
    {\tt\small \{lichengzhi,hhy63,pjian,bityangzhen,1120222204,guozhongbin\}@bit.edu.cn}
}

\begin{document}
\maketitle
\let\thefootnote\relax\footnotetext{$^{\dagger}$Corresponding author.}
\begin{abstract}
Large language models (LLMs) often generate self-contradictory outputs, which severely impacts their reliability and hinders their adoption in practical applications. In video-language models (Video-LLMs), this phenomenon recently draws the attention of researchers. Specifically, these models fail to provide logically consistent responses to rephrased questions based on their grounding outputs. However, the underlying causes of this phenomenon remain underexplored. In this work, we adopt an interpretability-driven approach to analyze, statistically summarize, and intervention the potential factors of the phenomenon. We find that one of the primary reasons for the inconsistency in responses lies in the inability of cross-modal attention heads to effectively distinguish video tokens across different timestamps. To address this, we propose an attention enhancement method called \textbf{T}emporally \textbf{C}onditioned \textbf{A}ttention \textbf{S}harpening (\textbf{TCAS}), which constructs an enhancement objective based on attention distinctions to enhance the model's temporal resolution capability, thereby improving its temporal understanding logic consistency. Experimental results demonstrate that our method significantly enhances the temporal logic consistency of Video-LLMs. Further analyses reveal that our method indeed improves the temporal discriminability of attention heads, validating our conclusions. Additionally, our method even achieves performance improvements in general video temporal grounding tasks, suggesting that temporal logic consistency is an important factor in temporal understanding. 
\end{abstract}    
\section{Introduction}
\label{sec:intro}

In recent years, Multimodal Large Language Models (MLLMs) have achieved significant strides in visual-language understanding~\cite{yue2023mmmu, pmlr-v235-ying24a}. Among them, Video Large Language Models (Video-LLMs) excel by combining video-based visual information with the pre-trained knowledge of large language models, enabling better comprehension and generation of natural language related to video content~\cite{vidllmsurvey}. These models have demonstrated outstanding performance in tasks like video question answering and captioning, driving progress in multimodal intelligence~\cite{videoMme}. However, the redundancy in video data challenges large language models in fine-grained temporal understanding. To address this, many studies aim to enhance the temporal understanding~\cite{guo2024vtgllm, timechat}, particularly for fine-grained tasks like Video Temporal Grounding (VTG).

These works often rely on additional specialized temporal modules to enhance Video-LLMs by providing additional temporal information streams~\cite{guo2024vtgllm, timechat, hu2024enhancing}. Jung et al.~\cite{consistency} proposed a benchmark to assess the consistency of state-of-the-art (SOTA) Video-LLMs, revealing their significant limitations in fine-grained temporal understanding. Specifically, Jung et al.~\cite{consistency} find that all Video-LLMs struggle to provide logically consistent responses to rephrased questions, highlighting a lack of true temporal comprehension in both the models and the enhancement methods.

Although many modular methods have been proposed to enhance temporal understanding, the underlying reasons for large language models' limitations in this area remain underexplored. Jung's findings~\cite{consistency} further reinforce the necessity for deeper analysis. Our work begins with temporal inconsistency, constructing an interpretability-driven analysis to uncover its intrinsic causes and deeply exploring temporal understanding. To our knowledge, this is the first study to analyze the temporal understanding consistency of Video-LLMs from an interpretability perspective. We focus on the following research questions:

\textit{1). What factors influence temporal understanding consistency in Video-LLMs, and how do these internal factors affect it?} In Section~\ref{sec:consistency_factors}, we use key head identifying, attention visualization, statistical summarization, and causal intervention methods, investigating the relationship between temporal understanding consistency and attention distributions. \textbf{Finally, our findings reveal and confirm that the primary cause of response inconsistency lies in the inability of cross-modal attention heads to effectively differentiate video tokens across different timestamps}.

\textit{2). How can we enhance the temporal understanding consistency of Video-LLMs based on the identified internal factors?} In Section~\ref{sec:attention_enhancement}, we introduce \textbf{T}emporally \textbf{C}onditioned \textbf{A}ttention \textbf{S}harpening (\textbf{TCAS}), an optimization objective based on attention distributions. This compels the cross-modal attention heads to explicitly judge the relevance of temporally distributed information, improving the model's temporal logic consistency. In Section~\ref{sec:experiments}, we validate the effectiveness across various experimental settings, demonstrating its ability to enhance temporal discriminability. Moreover, our method even improves performance in general VTG tasks, suggesting that temporal logic consistency is an important factor in temporal understanding.
\begin{figure}
\centering
\includegraphics[width=0.98\linewidth]{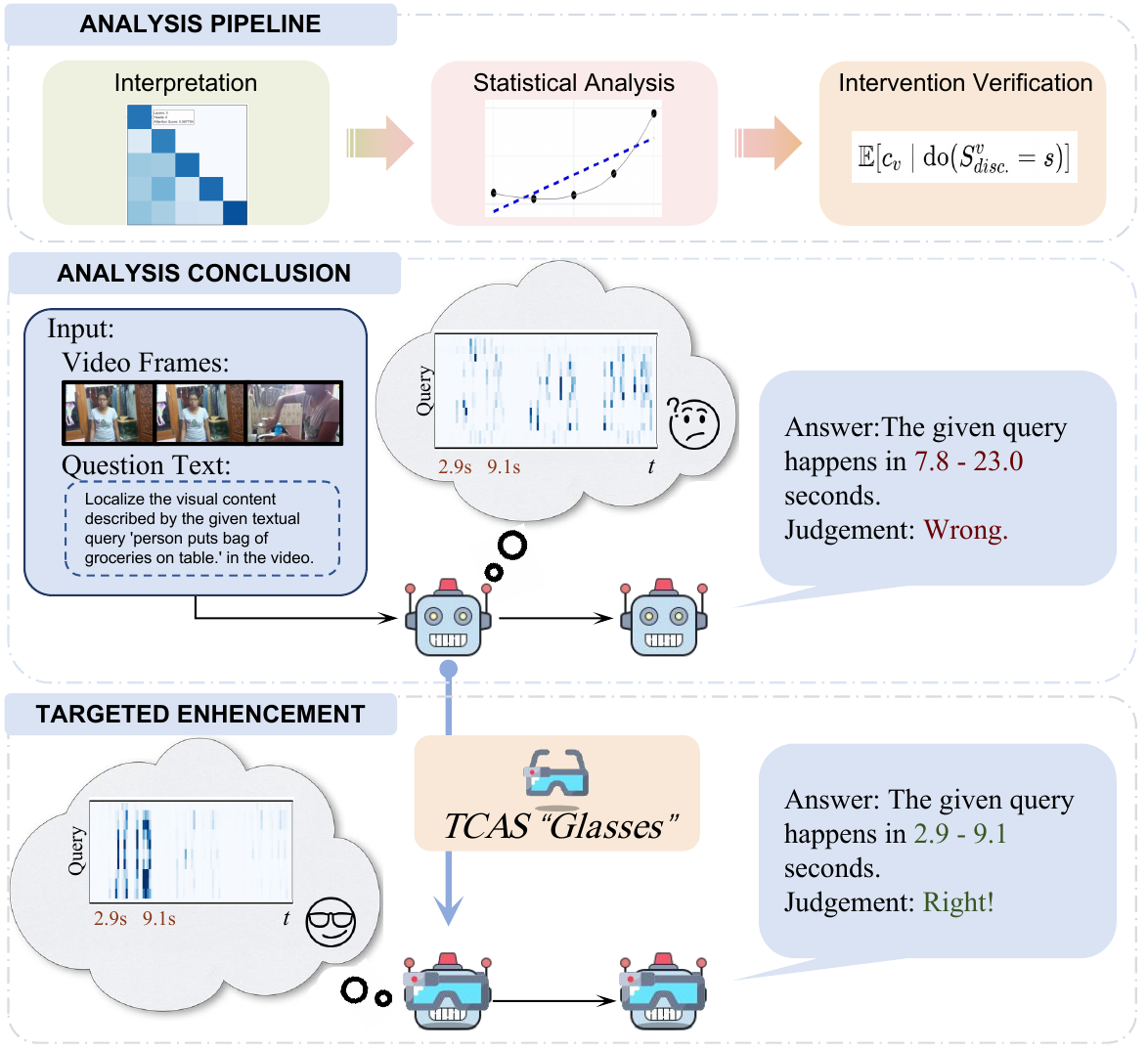}
\caption{Overview of our work. We first analyze to gain conclusions about the factors influencing temporal understanding logic consistency in Video-LLMs, and then propose a method to enhance the temporal consistency based on these conclusions.}
\label{1}
\end{figure}

The workflow is shown in Figure~\ref{1}. The contributions of this paper are summarized as follows:
\begin{itemize}
        \item We reveal the critical role of minority key cross-modal attention heads in temporal understanding, particularly in maintaining temporal understanding logic consistency.
    \item We conclude that the temporal discriminability of attention plays a crucial role in enhancing the model's prediction of temporal logic consistency.
    \item We propose TCAS, which improves temporal logic consistency by optimizing attention distributions, validated through extensive experiments.

\end{itemize}

\section{Related Work}
\subsection{Video-LLMs on VTG}

Recent Video-LLMs excel in fine-grained temporal understanding tasks like VTG by independently encoding timestamps and interacting with video information to capture precise temporal segments, marking a significant improvement over traditional VTG methods by enabling diverse natural language outputs~\cite{guo2024vtgllm, timechat, hu2024enhancing}. New benchmarks for fine-grained temporal understanding have been proposed to analyze their limitations~\cite{videoMme, chen2024rextime, consistency}. For example, the consistency benchmark by Jung et al.~\cite{consistency} highlights response inconsistency. Unlike prior work, we focus on the internal representations of Video-LLMs rather than solely developing new benchmarks for evaluation, providing deeper insights into their temporal cognition.

\subsection{Interpretability Analysis in MLLMs}

In recent years, interpretability of LLMs has emerged as a key research focus to enhance model transparency and controllability~\cite{rai2025practicalreviewmechanisticinterpretability}. However, the understanding of MLLMs remains limited~\cite{lin2025surveymechanisticinterpretabilitymultimodal}. 
Nikankin~\cite{nikankin2025task} showed that vision–language models rely on largely disjoint modality-specific circuits~\cite{elhage2021mathematical}, with visual representations aligning to textual ones only in later layers. Venhoff~\cite{venhoff2025visualrepresentationsmaplanguage} found that visual features gradually converge with language representations across layers, revealing an early-layer misalignment in adapter-based multimodal architectures.
More recently, Yu and Ananiadou~\cite{yu2025understandingmultimodalllmsmechanistic} conducted a mechanistic analysis of LLaVA and revealed that its visual reasoning process parallels the in-context learning dynamics of text-only LLMs.

The above work focuses on explaining how image concepts align with text representations. For temporal reasoning, Li et al.~\cite{li2025videot3} probed visual features and identified the LLM decoder as the bottleneck, but did not explore the root causes of this limitation. From a consistency perspective, we offer a deeper explanation, attributing the decoder’s limitations to the insufficient discriminability of cross-modal attention heads.
\section{Exploring Temporal Logic Consistency}

\label{sec:consistency_factors}

We conduct experiments using the original grounding (G.), rephrased grounding (R-G.), and shifted grounding (S-G.) subsets of Charades-CON~\cite{consistency}. For each sample, we define the model's \textbf{Rephrase Consistency Score} $c^{v}_{rg.}$ and \textbf{Shift Consistency Score} $c^{v}_{sg.}$ as $I^{v}_{ori} \times I^{v}_{rg.}$ and $I^{v}_{ori} \times I^{v}_{sg.}$, respectively, where $I^{v}_{ori}$, $I^{v}_{rg.}$, and $I^{v}_{sg.}$ represent the Intersection over Union (IoU) between the model predictions and the ground-truth timestamps for original, rephrased, and shifted grounding tasks.

This section focuses on analyzing the TimeChat-7b model~\cite{timechat}, which is a multimodal large language model used for long Video understanding and accurate time localization tasks, fusing temporal features and visual information through Sliding Video Q-Former and Timestation-aware frame encoder. In addition, for Generalizability of our conclusions, we analyze Qwen2.5-VL~\cite{qwenvl} on the Event Order Judgment (EOJ) task in Appendix~\ref{apd:generalizability_analysis}.

\begin{figure}[t]
    \centering
    \includegraphics[width=0.9\columnwidth,height=0.55\columnwidth]{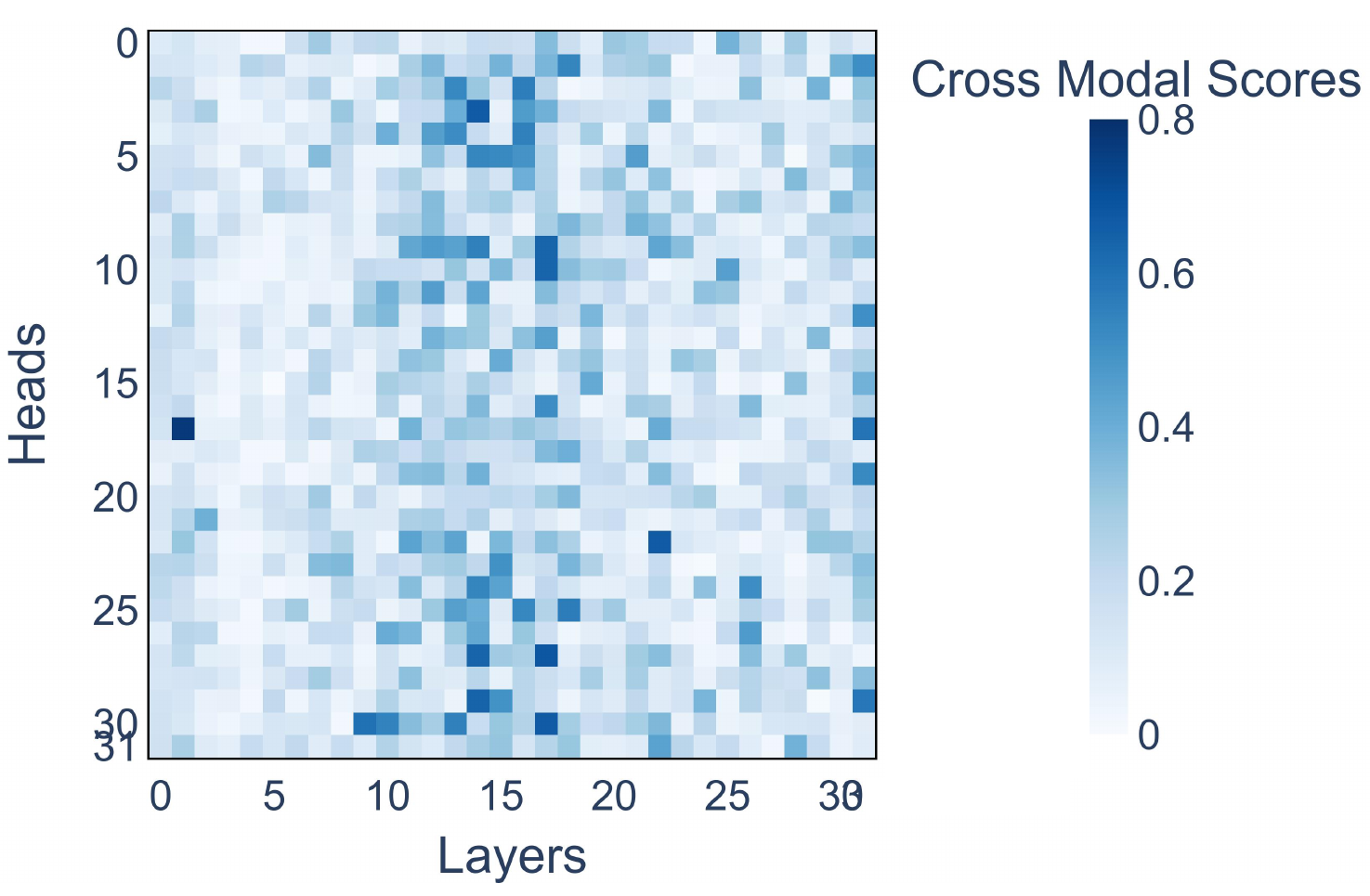}
    \caption{The distribution of cross-modal attention heads in TimeChat. The x-axis represents the attention layer index, and the y-axis represents the head index.}
    \label{fig:cross_modal_attention_heads}
\end{figure} 
\subsection{Detection and Interpretation}

TimeChat utilizes a Q-Former module to transform video features extracted by the Encoder into visual tokens, which are then fed into the large language model. Therefore, we start from these visual tokens to locate and visualize the attention heads that particularly focus on visual tokens, and explore how their behavior affects temporal logic consistency performance. For any attention head $h$, its \textbf{Cross-Modal Score} on video sample $v$ is defined as the average of the attention scores from all event text tokens to visual tokens, specifically defined as follows:

\begin{equation}
    S^{h,v}_{cross.} = \frac{1}{|T|} \sum_{q \in T} \sum_{k \in V} A^{h,v}_{q,k} \text{,}
    \label{eq:cross_modal_score}
\end{equation}
where $S^{h,v}_{cross.}$ denotes the cross-modal score, $T$ represents the tokens set of questions and answer text, $V$ represents the set of visual tokens, and $A^{h,v}_{q,k}$ is the attention score from text token as query token $q$ to visual token as key token $k$.

By sorting attention heads based on their cross-modal scores, we can identify those that particularly focus on visual tokens, which we refer to as cross-modal attention heads. The distribution of these cross-modal attention heads is shown in Figure~\ref{fig:cross_modal_attention_heads}. It is observed that only a small portion of mid-layer attention heads in the TimeChat model exhibit significantly higher cross-modal scores. These cross-modal attention heads align video information with text semantics, enabling the model to understand video content.
\textit{How do these cross-modal attention heads establish the model's temporal cognition?} 

\begin{figure*}[t]
    \centering
    \includegraphics[width=\textwidth]{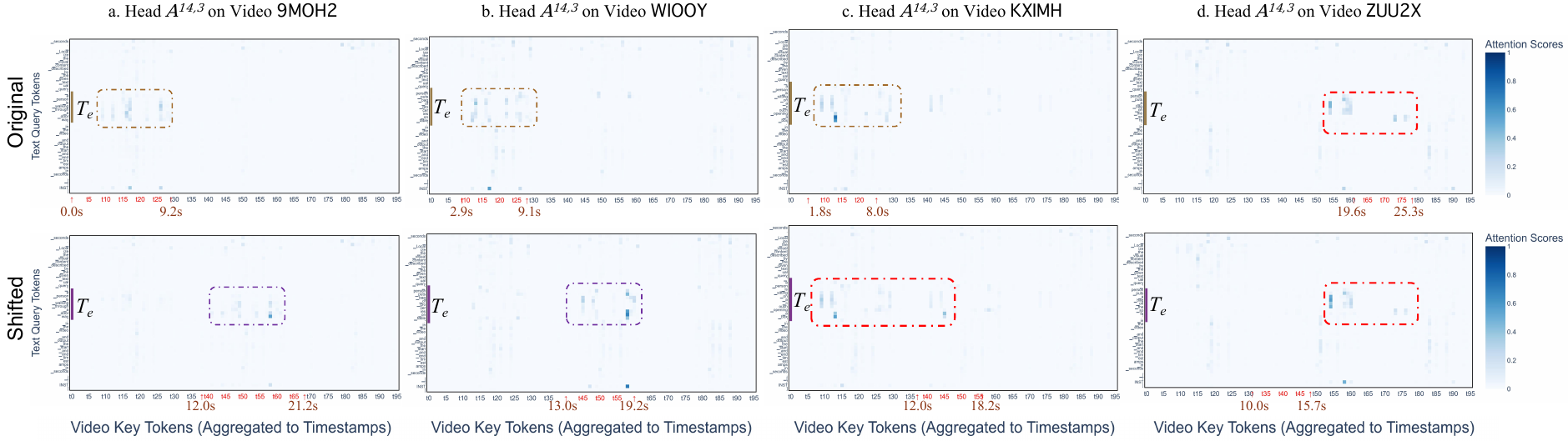}
    \caption{Visualization of attention score distributions for key head \( A^{14,3} \) across various samples. The start and end frames of the query event are marked in red at the corresponding token positions. }
    \label{fig:pattern_visualization}
\end{figure*}

To further investigate this question, we conduct a visual analysis of the behavior of cross-modal attention heads in the temporal dimension. The tokens in the key dimensions are aggregated based on their corresponding timestamps to highlight the attention head's discriminability in the temporal dimension while disregarding spatial dimensions. In Figure~\ref{fig:pattern_visualization}, we present the distribution of attention scores for high-score cross-modal attention head $A^{14,3}$ across various samples.

\begin{table}[t]
\caption{Prediction of samples for attention visualization.}
\label{tab:pattern_visualization}
\centering
  \fontsize{9pt}{9pt}\selectfont
\begin{tabular}{c|cccc}
\toprule
\textbf{IoU} & \textbf{9MOH2} & \textbf{WIOOY} & \textbf{KXIMH} & \textbf{ZUU2X} \\
\midrule
$I_{ori}$ & 0.956 & 0.451 & 0.413 & 0.414 \\
$I_{sg.}$ & 0.594 & 0.639 & 0 & 0 \\
$c_{sg.}$ & 0.562 & 0.288 & 0 & 0 \\
\bottomrule
\end{tabular}

\end{table}

Figure~\ref{fig:pattern_visualization} shows how key head $A^{14,3}$ maps event text tokens to corresponding temporal visual tokens, enabling cross-modal alignment. However, $A^{14,3}$ fails to focus on the segments corresponding to the events in videos KXIMH and ZUU2X, correlating with poor consistency performance (Table~\ref{tab:pattern_visualization}). This suggests that this property of cross-modal attention heads could be a critical factor influencing the model's temporal logic consistency. Additional patterns of other heads and samples can be found in Appendix~\ref{apd:attention_patterns}.

\subsection{Statistical Analysis}

We define \textbf{Attention Discriminability Score} as a measure of an attention head's ability to differentiate event time segments within a video sample. This definition allows us to analyze the relationship between attention head behavior and the model's consistency performance. Specifically, for any attention head $h$ and a video sample $v$ containing the query event $e$, attention discriminability is quantified using the ratio of the attention scores assigned to video tokens within the ground-truth time range to the total attention scores assigned to all video tokens.
The formal definition is as follows:
\begin{gather}
    S^{h,v}_{disc.} = \frac{1}{|T_{e}|} \times \sum_{q \in T_{e}} \frac{\sum_{k \in V_{gt}} A^{h,v}_{q,k}}{\sum_{k \in V} A^{h,v}_{q,k}} \text{,} \label{eq:attention_discriminability} \\
    S^{v}_{disc.} = \frac{1}{|H_{t}|} \sum_{h \in H_{t}} S^{h,v}_{disc.} \text{,}
    \label{eq:average_attention_discriminability}
\end{gather}
where $T_{e}, V_{gt}$ represent the set of text tokens of event $e$ and the set of visual tokens in the ground-truth time range for event $e$, respectively. $S^{h,v}_{disc.}$ is the temporal discriminability score of attention head $h$ for video sample $v$, $H_{t}$ means the set of the top $t$ cross-modal score attention heads, and $S^{v}_{disc.}$ is the average temporal discriminability score of the top $t$ cross-modal attention heads for video sample $v$.

\begin{figure}[t]
    \centering
    \includegraphics[width=0.95\columnwidth,height=0.85\columnwidth]{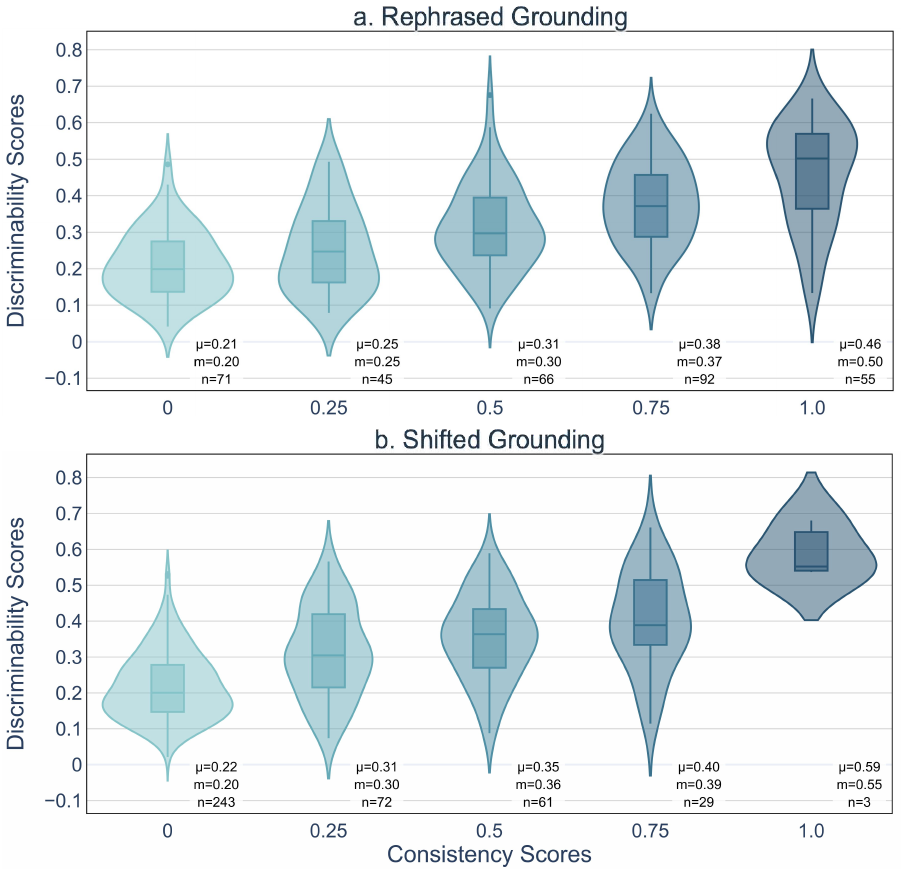}
    \caption{Violin plot showing the distribution of attention discriminability scores for different ranges of consistency scores. $\mu$, m, n denote the mean, median, and number of the discriminability score distribution, respectively.}
    \label{fig:corr}
\end{figure}

For each video sample $v$, we calculate its consistency scores $c^{v}_{rg.}$, $c^{v}_{sg.}$ and attention discriminability score $S^{v}_{disc.}$. The Pearson correlation coefficient is computed by aggregating the values of these two metrics across all samples in the Charades-CON dataset. The discriminability scores distribution across different consistency sets is shown in Figure~\ref{fig:corr} as a violin plot, where we can find that $S_{disc.}$ and $c^{v}$ are positively correlated.
The Pearson correlation coefficients are $0.4778$ (with p-value $\mathbf{5.18 \times 10^{-41}} \ll 0.05$) and $0.4788$ (with p-value $\mathbf{3.69 \times 10^{-41}} \ll 0.05$) for the rephrased grounding and the shifted grounding sets. This correlation coefficient is evidently statistically significant for Video-LLMs, which are inherently complex nonlinear systems. This result demonstrates a unambiguous positive correlation between the temporal discriminability of key heads and the model's consistency performance.
The scatter plot, the least-squares regression line, and more discussions are shown in Appendix~\ref{apd:correlation_analysis_details}.

\begin{table}[t]
\caption{Post-intervention scores with different $\alpha$ values. $\mathcal{M}$, R0.5, and R0.7 represent mIoU, recall at IoU thresholds of 0.5 and 0.7.}
\label{tab:intervened_consistency}
\centering
  \setlength{\tabcolsep}{2pt}
  \fontsize{9pt}{10pt}\selectfont

\begin{tabular}{cc|ccc|ccc|ccc}
\toprule
\multirow{2}{*}{$\alpha$} & \multirow{2}{*}{$S_{disc.}$} & \multicolumn{3}{c}{\textbf{G}}& \multicolumn{3}{c}{\textbf{R-G.}}& \multicolumn{3}{c}{\textbf{S-G.}} \\
& & {R0.5} & {R0.7} & {$\mathcal{M}$}& {R0.5} & {R0.7} & {$\mathcal{M}$}& {R0.5} & {R0.7} & {$\mathcal{M}$}\\
\midrule
w/o &0.30  & 75.3 & 57.5 & 65.8 
           & 90.2 & 79.9 & 81.8 
           & 45.8 & 24.0 & 41.4 \\ 
\hline
\rule{0pt}{1.0em}
0.2 &0.42  & \textbf{75.4} & 58.0 & 66.0 
           & \textbf{90.7} & \textbf{80.8} & \textbf{82.6} 
           & 46.4 & 22.8 & 41.8 \\
0.4 &0.54  & 75.4 & \textbf{58.5} & \textbf{66.0} 
           & 90.0 & 79.5 & 81.9 
           & 46.0 & \textbf{24.5} & 42.4 \\
0.6 &0.65  & 75.1 & 57.6 & 65.4 
           & 90.0 & 78.1 & 80.9 
           & 45.2 & 23.3 & 42.9 \\
0.8 &0.77  & 74.4 & 57.3 & 64.6 
           & 89.0 & 76.5 & 79.8 
           & 46.0 & 23.9 & 43.8 \\
1.0 &0.86  & 73.3 & 54.3 & 63.2 
           & {88.5} & {74.3} & {78.5} 
           & \textbf{47.7} & 23.5 & \textbf{44.2} \\

\bottomrule
\end{tabular}

\end{table}

\begin{figure*}[t]
    \centering
    \includegraphics[width=\textwidth]{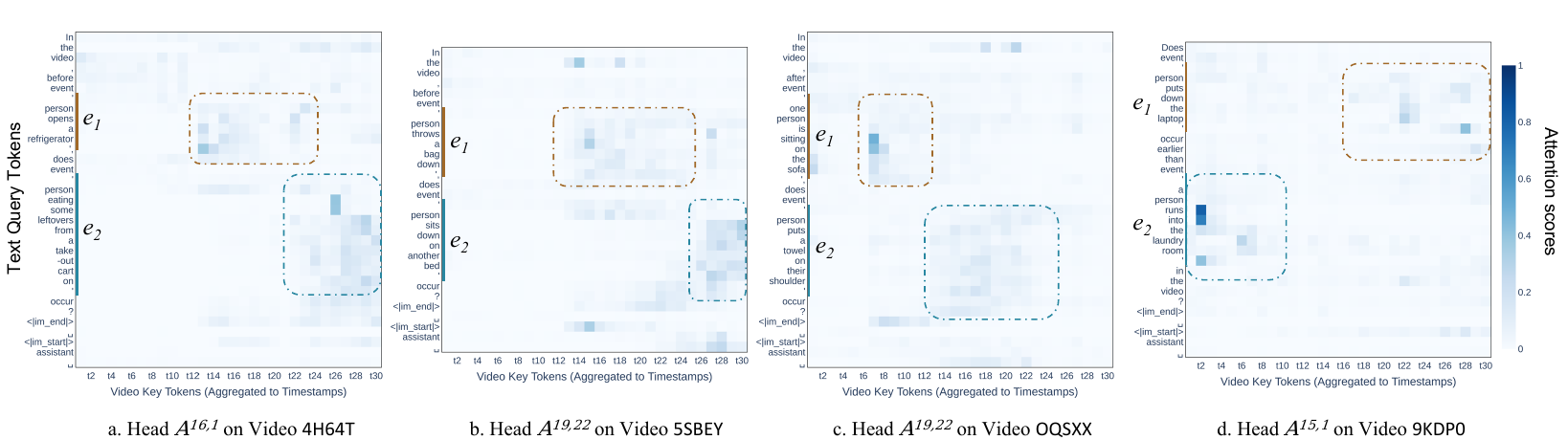}
    \caption{Attention patterns of key heads in the EOJ task.}
    \label{fig:attention_patterns_eoj}
\end{figure*}

{\bf Generalizability.} To briefly explore the generalizability across other models and tasks, we extend our observation to the Qwen2.5-VL~\cite{qwenvl} on the Event Order Judgment (EOJ) task. As illustrated in Figure~\ref{fig:attention_patterns_eoj}, we can observe that the text description tokens of events occurring at different times can generally establish appropriate focus on their corresponding temporal visual segments. This phenomenon implies the cross-model and cross-task generalizability of our findings. More detailed analysis is provided in Appendix~\ref{apd:generalizability_analysis}.

\subsection{Intervention Verification}

Statistical correlation alone is often insufficient to establish a causal relationship~\cite{Causality}. To further validate the causal relationship between attention discriminability and temporal logic consistency, a targeted intervention~\cite{Causality} study is conducted to determine whether improving attention discriminability leads to a rise in temporal logic consistency. Specifically, we intervene on attention for each query token $q \in T_e$ and head $h \in H_t$ via
\begin{equation}
A_{q,V} = (1-\alpha)A_{q,V}^{orig} + \alpha A_{q,V}^{gt}. 
    \label{eq:attention_intervention}
\end{equation}
Here, $V$ is video tokens set, $A^{orig}$ is the original attention, and $A^{gt}$ is the ground-truth attention, a uniform distribution over tokens within the event timestamps. Targeted minor intervention isolates the effect of $S_{disc}$ excluding confounding factors.
This process guides the model to focus more on the ground-truth temporal segments during inference. 

By evaluating the model's consistency performance after the intervention, we aim to confirm the causal relationship between attention discriminability and temporal logic consistency.

The post-intervention discriminability scores and consistency scores are presented in Table~\ref{tab:intervened_consistency}. 
As shown in the table, mild interventions effectively improve the model's performance and consistency across the three subsets, while overly strong interventions lead to performance degradation. Additionally, relatively mild interventions are sufficient to enhance the model's $c_{rg.}$, whereas improving $c_{sg.}$ requires stronger intervention intensity, indicating that the shifted grounding task poses greater challenges for the model in maintaining consistent outputs. These results demonstrate a clear causal relationship between attention discriminability and temporal logic consistency.

\section{Attention Enhancement}
\label{sec:attention_enhancement}

In the previous section, it is confirmed that the temporal discriminability of cross-modal attention heads is a key factor influencing the model's temporal logic consistency. \textit{How can we enhance attention discriminability to improve temporal consistency?} To achieve generalization and cross-architecture applicability, we formulate the attention discriminability enhancement as an optimization objective termed \textbf{T}emporally \textbf{C}onditioned \textbf{A}ttention \textbf{S}harpening Loss (\textbf{TCAS} Loss). 

To ensure cross-task generalization across various video-language tasks, Our TCAS Loss does not rely on ground-truth temporal labels to guide attention distribution optimization. This loss function employs a contrastive learning approach, compelling the model to make explicit relevance judgments for information at different timestamps, thereby enhancing the discriminability of attention heads in the temporal dimension.

\begin{algorithm}[t]
\caption{TCAS Loss}
\label{alg:attention_enhancement}
\textbf{Input}: Video and text tokens $V,T$, model $M$ \\
\textbf{Parameter}: Top heads number $t$, margin $m$, threshold $thr$ \\
\textbf{Output}: TCAS loss $\mathcal{L}$
\begin{algorithmic}[1]
\setlength{\baselineskip}{1.2\baselineskip} 
\STATE Response $r$, attentions $A^{H} \gets M(V, T)$.
\STATE Compute $S^{v}_{cross.}$ using Equation~(\ref{eq:cross_modal_score}).
\STATE $H_{t} \gets \textsc{SelectTopHeads}(H,S^{v}_{cross.}, t)$ 

\FOR{each head $h$ in $H_{t}$}

    \STATE Aggregate visual tokens in $A^{h}_{T,V}$ according to timestamps to get $A^{h}_{T,V_{time}}$     \STATE $T^{h}_{valid} \gets \{q \in T \mid \text{Max}(A^{h}_{q,V_{time}}) > thr\}$

    \FOR {each text query token $q$ in $T^{h}_{valid}$}

        \STATE Mean $M^{h}_{q} = \sum_{k \in V_{time}}(A^{h}_{q,k}) / |V_{time}|$
        \STATE Pos. $P^{h}_{q} \gets \{A^{h}_{q, k} \in A^{h}_{q, V_{time}} \mid A^{h}_{q, k} > M^{h}_{q} \}$ 
        \STATE Neg. $N^{h}_{q} \gets \{A^{h}_{q, k} \in A^{h}_{q, V_{time}} \mid A^{h}_{q, k} < M^{h}_{q} \}$ 
        \STATE Loss $\mathcal{L}^{h}_{q} = \text{Max}(m + \text{Max}(N^{h}_{q}) - \text{Min}(P^{h}_{q}), 0)$
    \ENDFOR
\ENDFOR
\RETURN $ (\sum_{h \in H_{t}} \sum_{q \in T^{h}_{valid}} \mathcal{L}^{h}_{q}) / (\sum_{h \in H_{t}} |T^{h}_{valid}|)$ 
\end{algorithmic}
\end{algorithm}

\begin{table*}[t]
\caption{Comparison results for consistency and grounding performance. For the grounding task, IoU scores are reported, while accuracy rates are provided for the Verify task. Each cell shows the absolute performance, with relative performance in parentheses. Qwen2.5vl and videollama refers to Qwen2.5-VL-7B-Instruct and Video-LLaMA-7B.}
\label{tab:comparison_results}
\centering
\setlength{\tabcolsep}{1.2pt}
\fontsize{9pt}{10pt}\selectfont
{    
    \begin{tabular}{ccccccccccccc}
    \toprule
    \multirow{3}{*}{\bf {Method}} & \multirow{3}{*}{\bf Data} & \multirow{3}{*}{\bf FT} & \multicolumn{5}{c}{\bf Charades-CON} & \multicolumn{5}{c}{\bf ActivityNet-CON} \\
    \cmidrule(lr){4-8} \cmidrule(lr){9-13}
    & & & Ground & R-Ground & S-Ground & H-Verify & C-Verify 
    & Ground & R-Ground & S-Ground & H-Verify & C-Verify \\
                        \midrule

    Qwen2.5vl & \textit{VTune} & SFT
        
        & 28.3 & {17.5} (62.0) & 6.0 (21.1) & 15.1 (53.3) & 14.8 (52.1) 
    & 18.2 & {14.5} (79.8) & 2.5 (13.5) & 12.0 (66.0) & 13.2 (72.6) \\
                                \rowcolor{lightgray}
    Qwen2.5vl & \textit{VTune} & TCAS
    & \underline{34.0} & \underline{23.0} (67.5) & \underline{8.1} (23.7)  & \underline{19.6} (57.6) & \underline{18.5} (54.3)
    & \underline{18.7} & \underline{15.1} (80.5) & {2.5} (13.6) & \underline{12.3} (65.5) & \underline{14.0} (74.6) \\
    \midrule

    videollama & \textit{VTune} & SFT
    & 54.4 & 38.2 (70.3) & 10.9 (20.0)  & 30.7 (56.5) & 30.0 (55.2)
    & 33.0 & 24.7 (74.8) & 10.0 (30.2) & 20.2 (61.1) & 17.7 (53.7) \\

    \rowcolor{lightgray}
    videollama & \textit{VTune} & TCAS
    & \underline{62.4} & \underline{44.4} (71.2) & \underline{13.4} (21.4)  & \underline{34.8} (55.8) & \underline{37.3} (59.7)
    & \underline{33.8} & \underline{25.2} (74.7) & \underline{11.1} (32.7) & \underline{20.9} (61.8) & \underline{18.6} (55.1) \\
    \midrule

        TimeChat & \xmark & \xmark 
    & 30.5 & 25.0 (82.1) & 5.6 (18.5) & 14.0 (45.9) & 15.6 (51.2)
    & 4.6 & 2.9 (64.1) & 1.0 (21.2) & 2.1 (46.7) & 2.4 (52.2) \\
    
    TimeChat & \textit{TimeIT} & SFT
    & {{55.8}} & {{50.9}} (91.3) & {{10.5}} (18.9) & {{16.7}} (30.0) & {{25.7}} (46.2)
    & {{25.3}} & {{20.2}} (80.4) & {{7.5}} (29.9) & {{8.7}} (34.5) & {{12.6}} (49.9) \\

    \rowcolor{lightgray}
    TimeChat & \textit{TimeIT} & TCAS
    & {66.2} & {60.7} (91.3) & {33.8} (51.1) & {37.3} (56.3) & {31.4} (47.4)
    & {38.1} & {29.2} (76.6) & {10.4} (27.3) & \textbf{24.2} (63.3) & {18.3} (47.9) \\
    
    TimeChat & \textit{VTune} & SFT
    & { 76.2} & {69.2} (90.8) & {36.2} (47.5) & {44.8} (58.8) & {42.4} (55.7)
    & {37.4} & {28.3} (75.6) & {10.6} (28.3) & {19.6} (52.3) & {19.3} (51.5) \\

    \rowcolor{lightgray}
    TimeChat & \textit{VTune} & TCAS
    & {\bf 83.3} & {\bf 75.0} (90.1) & {\bf 39.5} (47.4) & {\bf 52.9} (63.5) & {\bf 50.8} (61.0)
    & {\bf 38.8} & {\bf 31.0} (79.9) & {\bf 11.1} (28.6) & {{20.8}} (53.6) & {\bf 21.5} (55.5) \\
    \bottomrule
    \end{tabular}
}

\end{table*}

The specific loss computation is detailed in Algorithm~\ref{alg:attention_enhancement}, where the $\textsc{SelectTopHeads}(H, S, t)$ means selecting the top $t$ attention heads from the set of all heads $H$ based on scores $S$. 
We first identify the attention heads with the highest cross-modal scores and the text tokens that focus on specific video timestamps. This approach preserves the original contextual interaction capabilities of the attention heads. A contrastive learning loss is designed for the selected text query tokens to amplify the differences in attention scores across distinct timestamps, thereby compelling the attention heads to make explicit relevance judgments for temporally distributed information. We use weight $w_{ae}$ to balance the TCAS loss and next token prediction loss. Our method introduces no additional modules, ensuring compatibility with various transformer-based architectures~\cite{transformer}. Furthermore, it does not depend on specific text templates or video patterns, making it broadly applicable to diverse video-language tasks.

\section{Experiments}
\label{sec:experiments}
In this section, we present a series of experiments to evaluate the effectiveness of our proposed method in enhancing grounding performance and logic consistency in the VTG task. This section describes the experimental setup, presents comparison results, analyzes component contributions, provides some prediction examples, and reports the attention discriminability scores after enhancement.

\subsection{Experimental Setup}
\paragraph{Models}
We conduct experiments on the Qwen2.5-VL~\cite{qwenvl}, Video-LLaMA~\cite{videollama}, and TimeChat~\cite{timechat} models, which respectively represent Video-LLMs based on MLP projection, Video-LLMs based on Qformer, and time-aware models. More details about these models can be found in Appendix~\ref{apd:backbone_models}.

\paragraph{Datasets}
We conduct experiments on two widely-used VTG datasets: \textbf{ActivityNet Captions}~\cite{ActivityNet} and \textbf{Charades}~\cite{charadesSTA}. ActivityNet Captions comprises approximately 20k untrimmed YouTube videos with an average duration of 120 seconds, while Charades contains 9848 short videos with an average length of 30 seconds. Additionally, \textbf{Charades-CON} and \textbf{ActivityNet-CON}, introduced by Jung et al.~\cite{consistency}, serve as consistency benchmarks to ensure fair, transparent, and reliable comparisons.

\paragraph{Tuning Annotations}

We select two different training Annotations. The first is a standard instruct annotation dataset for the VTG consistency task, called \textit{TimeIT}, proposed by Ren et al.~\cite{timechat}. The second dataset is \textit{VTune}, introduced by Jung et al.~\cite{consistency}, which has been shown to effectively enhance the consistency of Video-LLMs.
Our method demonstrates improved results over baseline fine-tuning methods using both datasets, indicating that it does not rely on specific data sources. The dataset examples can be found in Appendix~\ref{apd:datasets_example}.

\paragraph{Evaluation Metrics}
Following previous works~\cite{consistency,timechat}, we evaluate grounding performance using IoU metrics, while measuring the consistency of the model responses through the accuracy rate. Details about the evaluation metrics can be found in Appendix~\ref{apd:evaluation_metrics}.

\paragraph{Implementation Details}
We implement our method using PyTorch on a single NVIDIA A100 80GB GPU. All settings are trained with the Adam optimizer (learning rate: $1 \times 10^{-5}$, batch size: 4) for approximately 3 days. Qwen2.5-VL, TimeChat, and Video-LLaMA process 16, 96, and 8 frames respectively, all at $224 \times 224$ spatial resolution.

\begin{table}[t]
\caption{Comparison results for VTG performance on Charades-STA and ActivityNet-Caption. R@1,0.5 and R@1,0.7 represent Recall at IoU thresholds of 0.5 and 0.7.}
\label{tab:comparison_results_VTG}
\centering
\setlength{\tabcolsep}{2pt}
\fontsize{9pt}{10pt}\selectfont
      \begin{tabular}{lcccc}
      \toprule
      \multirow{3}{*}{\textbf{Method}} & \multicolumn{2}{c}{\bf Charades-STA} & \multicolumn{2}{c}{\bf ActivityNet-Cap.} \\
      \cmidrule(l{3pt}r{3pt}){2-3} \cmidrule(l{3pt}r{2pt}){4-5}
       & R@1,0.5 & R@1,0.7 & R@1,0.5 & R@1,0.7 \\ \midrule
      {\textit{Task-Specific Models}} \\
              BM-DETR~\cite{Jung_2025_WACV} 
       &\textit{59.4} &\textit{38.3} &\textit{49.6} &\textit{30.6} \\ 
              Mr.BLIP~\cite{meinardus2024chrono} 
       &\textit{\underline{69.3}} &\textit{\underline{49.2}} &\textit{\underline{53.9}} &\textit{\underline{35.5}} \\ \midrule

      {\textit{Video-LLMs}} \\
       HawkEye~\cite{wang2024hawkeye} & \underline{58.3} & 28.8 & \underline{34.7} & 17.7 \\
       VTG-LLM~\cite{guo2024vtgllm} & 57.2 & \underline{33.4} & - & - \\

        TimeChat~\cite{timechat} & 46.7 & 23.7 & 28.0 & 15.8 \\
        \midrule

        TimeChat (w. SFT) & 58.4 & 34.7 & 41.0 & 23.7 \\

        \rowcolor{lightgray}
        \bf TimeChat (w. TCAS) & \textbf{60.2} & \textbf{37.6} & \textbf{41.2} & \textbf{24.9} \\
      \bottomrule
      \end{tabular}

\end{table}

\subsection{Comparison Experiments}
We conducted experiments to evaluate our method's grounding performance and consistency, demonstrating the effectiveness of our method, TCAS, in improving temporal logic consistency and understanding.

\subsubsection{Baselines}

We evaluate our method's effectiveness by comparing it with several baseline models and approaches. For open-source models, direct comparison without fine-tuning would be unfair, as no existing modeling methods specifically target consistency enhancement. Therefore, we compare our approach with standard fine-tuning methods, specifically \textbf{S}upervised \textbf{F}ine-\textbf{T}uning (SFT)~\cite{zhang2024instructiontuninglargelanguage}, applied to several selected backbones. The results of other models or backbones are listed in Appendix~\ref{apd:Other_Baselines}.

\subsubsection{Main Results}

We present the comparison results on consistency (Table~\ref{tab:comparison_results}) and grounding performance (Table~\ref{tab:comparison_results_VTG}). The best results are in bold or underlined in different areas. Light gray rows indicate results from our method, TCAS.
\begin{figure*}[t]
\centering
\includegraphics[width=\linewidth]{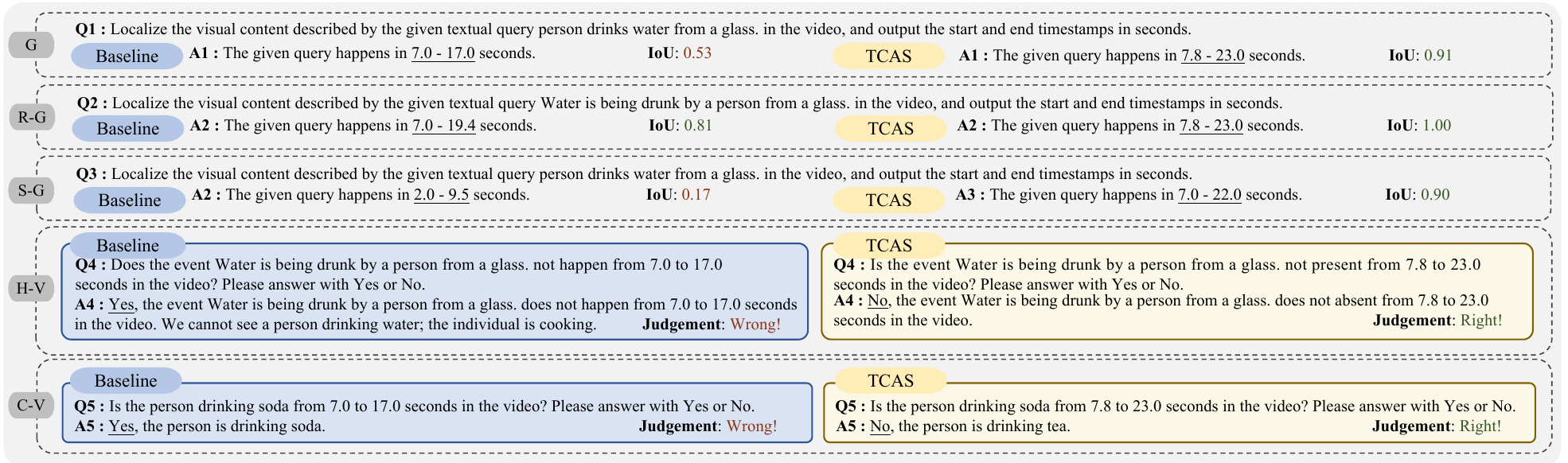}
\caption{Responses of TimeChat on grounding and verification for sample EQ4GN before and after TCAS enhancement.}
\label{fig:case_study}
\end{figure*}
\subsubsection{Discussion}

We focus on the following phenomena:

\textit{1). The improvements brought by TCAS are evident across different models, training datasets, and benchmarks.} As shown in Table~\ref{tab:comparison_results}, TCAS achieves favorable results in all comparisons, indicating that TCAS is not dependent on specific models, benchmarks, or training annotations and can robustly enhance model consistency. The performance improvement on ActivityNet-CON is relatively smaller, which is due to the noisier data distribution of this dataset, where event descriptions are presented in long sentences that deviate from the general understanding of the concept of events. Consequently, all methods and models perform worse on this dataset, making the smaller improvement acceptable.

\textit{2). TCAS not only improves consistency as anticipated but also yields unexpected gains in grounding performance.} This suggests that inconsistent temporal understanding is a potential factor limiting the grounding of Video-LLMs. We attribute this improvement to the regularization effect of TCAS. While standard SFT often fits dataset biases to satisfy token-level objectives—relying on shortcuts like \textit{language priors}~\cite{zheng2025mllmsdeeplyaffectedmodality,Zhang2024DebiasingML} rather than true grounding evidence—TCAS effectively suppresses such non-essential cues. By forcing the model to utilize discriminative attention on temporal tokens for reasoning, our method highlights authentic temporal evidence and improves generalization, ultimately advancing the model's fundamental temporal comprehension.

\begin{table}[t]
\caption{Ablation study results for hyperparameter sensitivity analysis. The table shows the absolute consistency scores for different hyperparameter configurations. }
\label{tab:ablation_results}
\centering
\setlength{\tabcolsep}{4pt}
\fontsize{9pt}{10pt}\selectfont
{    
    \begin{tabular}{ccccccccc}
    \toprule
    \multicolumn{4}{c}{\bf Hyperparameters}& \multicolumn{5}{c}{\bf Charades-CON}\\
    \cmidrule(l{4pt}r{3pt}){1-4} \cmidrule(l{4pt}){5-9} 
    ${t}$ & $m$ & $thr$ & $w_{ae}$ & G. & R-G. & S-G. & H-V. & C-V.\\
    \midrule

    \rowcolor{lightgray}
    \multicolumn{9}{c}{Analysis of top heads number $t$} \\
                                            16 & 0.2 & 0.1 & 0.5 
    & 80.91 & 72.14 & 36.77 & 40.98 & 42.17 \\

    48 & 0.2 & 0.1 & 0.5
    & 77.37  & 69.66  & 39.04 & 45.04 & 44.24\\

    \midrule
    \rowcolor{lightgray}
    \multicolumn{9}{c}{Analysis of margin $m$} \\
                                32 & 0.15 & 0.1 & 0.5 
    & 80.76 & 71.61 & 36.59 & 51.23 & 48.26 \\

                                        32 & 0.25 & 0.1 & 0.5
    & 78.64 & 71.07 & 34.35 & 49.48 & 45.09 \\

    \midrule
    \rowcolor{lightgray}
    \multicolumn{9}{c}{Analysis of threshold $thr$} \\

                                32 & 0.2 & 0.05 & 0.5 
    & \underline{81.90} & \underline{74.95} & \bf{41.45} & \bf{54.30} & \underline{48.95} \\

                                        32 & 0.2 & 0.15 & 0.5 
    &  81.05 & 72.33 & 34.84 & 44.10 & 40.65 \\

    \midrule
    \rowcolor{lightgray}
    \multicolumn{9}{c}{Analysis of loss weight $w_{ae}$} \\

    32 & 0.2 & 0.1 & 0.3 
    & 80.91 & 74.76 & 39.04 & 48.67 & 48.75\\

        32 & 0.2 & 0.1 & 0.7
    & 80.48 & 71.10 & 35.55 & 48.69 & 46.59 \\

    \midrule
    \rowcolor{lightgray}
    \multicolumn{9}{c}{Empirical optimal combination} \\
                                32 & 0.2 & 0.1 & 0.5
    & {\bf 83.31} & {\bf 75.02}& \underline{39.52} & \underline{52.93} & {\bf 50.81}\\

            \bottomrule
    \end{tabular}
}

\end{table}
\subsection{Ablation Study}

Our method is not composed of distinct independent modules, making traditional ablation studies infeasible. However, it depends on several key hyperparameters (i.e., top heads number $t$, margin $m$, threshold $thr$, and loss weight $w_{ae}$) that regulate the intensity of attention adjustment from various perspectives. Sensitivity analysis of these hyperparameters allows us to assess their contributions to improving temporal logic consistency and to further examine the intrinsic effectiveness of our approach.

The impact of different hyperparameters on model performance varies. The parameters $t$ and $thr$ primarily determine the adjustment scope of TCAS on the aspect of heads and tokens, respectively, while $m$ and $w_{ae}$ control the intensity of TCAS enhancement. As shown in Table~\ref{tab:ablation_results}, increasing $t$ leads to a significant drop in G. and R-G. subsets, while a larger $thr$ results in a substantial decline in S-G., H-V., and C-V. subsets, indicating that these two parameters are most critical for model performance. Overall, our method is more sensitive to scope-related hyperparameters while maintaining robustness to intensity hyperparameters.

\subsection{Case Study}
To visually demonstrate the effectiveness of our method, we select several prediction results from Charades-CON~\cite{consistency} for a case study. For the example shown in Figure~\ref{fig:case_study}, our model successfully grounds query events in the video and correctly answers the questions, demonstrating improved temporal reasoning capabilities compared to baseline methods. This result highlights the potential of our approach to enhance the understanding of temporal dynamics in video content. More examples are provided in the Appendix~\ref{apd:case_study}.

\subsection{Robustness to Video Duration}
\label{sec:video_length_analysis}
We evaluate the robustness of TCAS across different video durations on Charades-CON. As shown in Table~\ref{tab:video_length_analysis}, while the baseline performance typically declines in longer videos due to increased temporal complexity, TCAS yields progressively larger gains. Specifically, for videos exceeding 40 seconds, TCAS achieves substantial improvements of +17.7 points and +14.8 points in Grounding (G.) and Rephrase-Grounding (R-G.), respectively. This highlights its efficacy in maintaining discriminative attention and temporal consistency in challenging, long-form video understanding.

\begin{table}[t]
\caption{Robustness analysis across video durations on Charades-CON. $n$ is the sample count. Gains over SFT are shown in parentheses.}
\label{tab:video_length_analysis}
\centering
  \setlength{\tabcolsep}{2pt}
  \fontsize{9pt}{9pt}\selectfont
\begin{tabular}{c|cc|cc|cc|cc}
\toprule
& \multicolumn{2}{c}{\textbf{0--20s}, n58} & \multicolumn{2}{c}{\textbf{20--30s}, n189} & \multicolumn{2}{c}{\textbf{30--40s}, n409}& \multicolumn{2}{c}{\textbf{$\mathbf{>}$40s}, n51}\\
\cmidrule(lr){2-3} \cmidrule(lr){4-5} \cmidrule(lr){6-7} \cmidrule(l){8-9}
& SFT & TCAS & SFT & TCAS & SFT & TCAS & SFT & TCAS \\
\hline
\rule{0pt}{1.0em}
 G. & 89.7 & 93.1\scriptsize{\textcolor{delta}{+3.4}} & 78.8 & 84.1\scriptsize{\textcolor{delta}{+5.3}} & 74.1 & 82.4\scriptsize{\textcolor{delta}{+8.3}} & 58.8 & 76.5\scriptsize{\textcolor{delta}{+17.7}} \\
 R-G. & 84.5 & 86.2\scriptsize{\textcolor{delta}{+1.7}} & 72.3 & 76.4\scriptsize{\textcolor{delta}{+4.1}} & 66.3 & 73.9\scriptsize{\textcolor{delta}{+7.6}} & 51.2 & 66.0\scriptsize{\textcolor{delta}{+14.8}} \\
 S-G. & 46.6 & 55.2\scriptsize{\textcolor{delta}{+8.6}} & 35.5 & 38.1\scriptsize{\textcolor{delta}{+2.6}} & 34.1 & 39.7\scriptsize{\textcolor{delta}{+5.6}} & 20.9 & 25.5\scriptsize{\textcolor{delta}{+4.6}} \\
\bottomrule
\end{tabular}
\end{table}

\subsection{Visualization Analysis}
\label{sec:interpretability_analysis}

\begin{figure}[t]
\centering
\includegraphics[width=0.95\columnwidth]{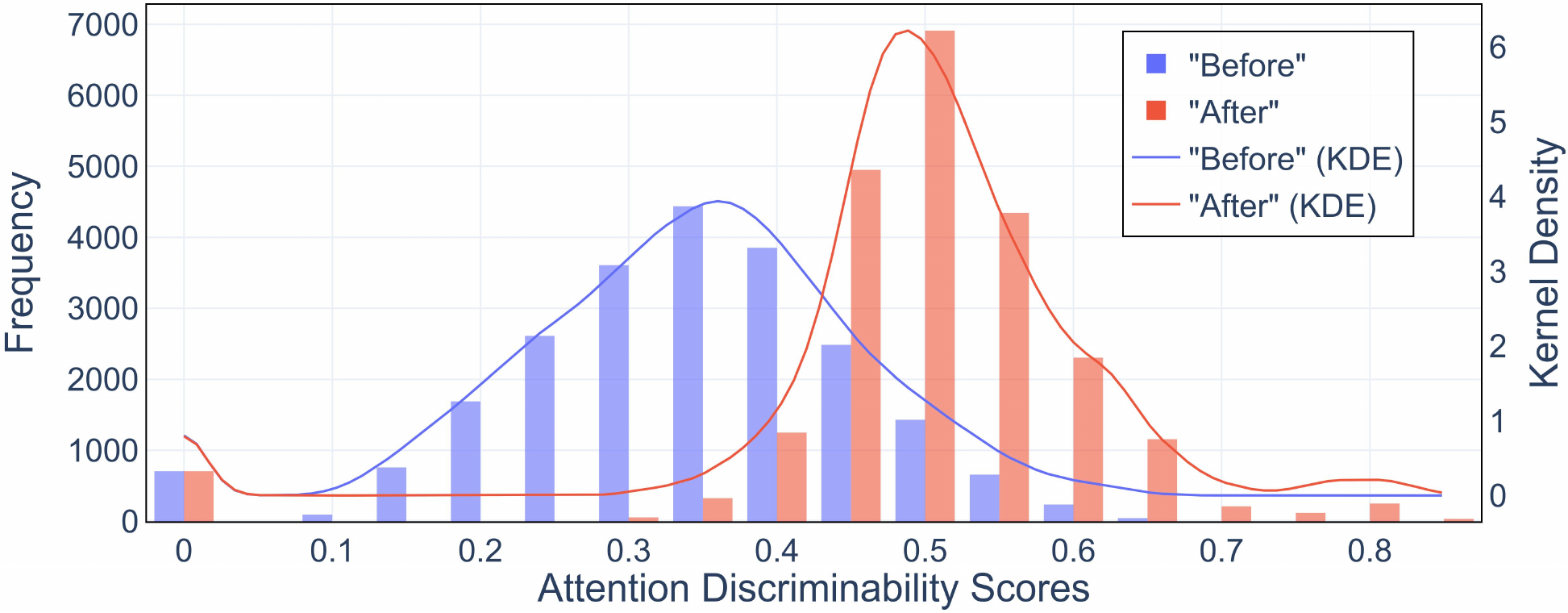}
\caption{Discriminability scores distribution \textquotesingle Before\textquotesingle{} (baseline model) and \textquotesingle After\textquotesingle{} (TCAS-enhanced model) attention enhancement. The KDE means Kernel Density Estimation curves.}
\label{fig:discriminability_scores_before_after}
\end{figure}

We analyzed the top-32 attention discriminability scores to confirm that TCAS's consistency improvements stem from enhanced discriminability, rather than unintended artifacts. Figure~\ref{fig:discriminability_scores_before_after} shows a clear rightward shift in the score distribution for the TCAS-enhanced model compared to the baseline. This result demonstrates that TCAS directly boosts attention discriminability, which consequently improves consistency, thus validating our approach, thereby reinforcing the validity of our conclusions.

\section{Conclusion and Limitation}

In this paper, we identify and validate a causal link between cross-modal attention discriminability and temporal logic consistency in Video-LLMs through correlation analysis and targeted intervention. Motivated by this finding, we introduce TCAS, a method that optimizes attention distributions to enhance temporal discriminability. Experiments demonstrate that TCAS significantly improves temporal logic consistency across various models. We acknowledge that our focus on logical inconsistency may not encompass all facets of temporal understanding.
\section*{Acknowledgement}

This work was supported by the National Natural Science Foundation of China (No. U21B2009) and the Foundation Program of the Key Laboratory of Science and Technology on Complex Electronic System Simulation (No. 614201001032203).

{
    \small
    \bibliographystyle{ieeenat_fullname}
    \bibliography{main}
}

\clearpage
\appendix
\setcounter{page}{1}
\maketitlesupplementary

In this file, we provide additional details as follows:
\begin{enumerate}[label=\Alph*.]     \item More Details about Analysis
    \begin{enumerate}[label=\arabic*.]                 \item \textbf{More Attention Patterns.} This section presents additional attention patterns of key attention heads in TimeChat, further supporting the analysis in Section 2.1 of the main text.
                \item \textbf{Correlation Analysis Details.} This section provides the fitting curves and complete distributions of the correlation analysis, as well as further analysis.

        \item \textbf{Details of Analysis on EOJ Task.} This section provides additional details on the analysis of the Event Order Judgement (EOJ) task, including the methodology, implementation details, results, and discussion.
    \end{enumerate}
    \item More Details about Experiments
    \begin{enumerate}[label=\arabic*.]
        \item \textbf{Backbone Models Introduction.} This section provides a detailed introduction to the backbone models used in our experiments, along with the reasons for their selection.
        \item \textbf{Datasets Introduction and Examples.} This section describes the datasets used in our experiments, including examples to illustrate their content and structure.
        \item \textbf{Evaluation Metrics.} This section details the evaluation metrics used to assess the performance.
        \item \textbf{Other Baselines and Backbones.} This section discusses additional baselines and backbone models that are considered in our experiments.
        \item \textbf{Case Study.} This section presents some case studies to illustrate the effectiveness of our method in enhancing temporal consistency in Video-LLMs.
        \item \textbf{Further Ablation Studies on Sensitivity.} This section provides additional ablation studies to further validate the components of our proposed method.
    \end{enumerate}
    \item More Discussions
    \begin{enumerate}[label=\arabic*.]
        \item \textbf{Method Design Related Work.} This section reviews related work on contrastive learning relevant to our method design.
        \item \textbf{Limitations and Future Work.} This section discusses the limitations of our current work and outlines potential directions for future research.
    \end{enumerate}
\end{enumerate}

\section{More Details about Analysis}

\subsection{More Attention Patterns}
\label{apd:attention_patterns}

\begin{figure*}[t]
    \centering
    \includegraphics[width=0.9\textwidth, height=0.60\textwidth]{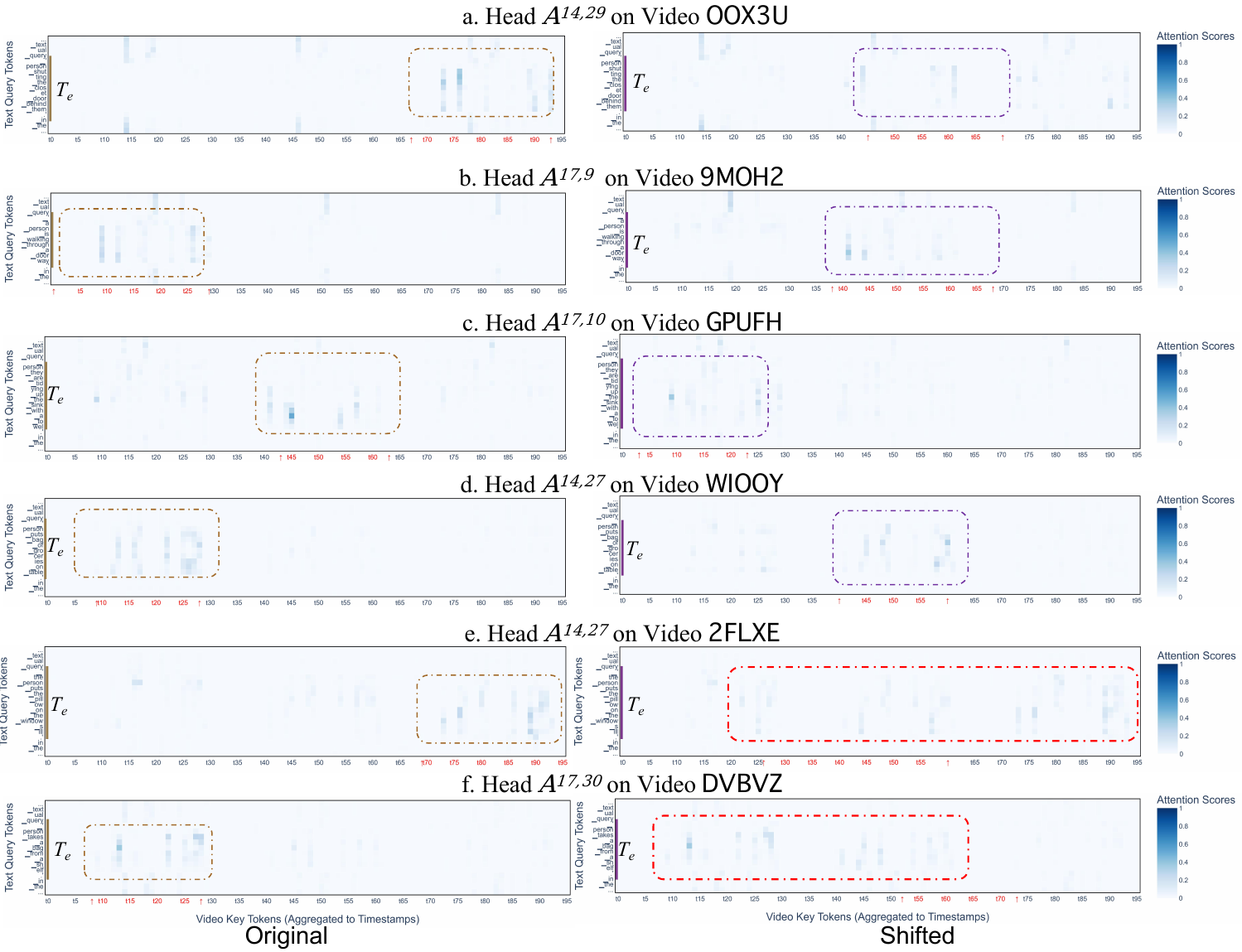}
    \caption{Visualization of attention score distributions for key heads across various samples. Subplots a-d are high discriminability cases, while e and f represent cases where the key heads fail to distinguish correctly.}
            \label{fig:pattern_visualization_apd}
\end{figure*}

In the main text, we presented the attention patterns of key attention heads in TimeChat. To further support the analysis in Section 2.1, we provide additional attention patterns of key attention heads in TimeChat.
As shown in Figure~\ref{fig:pattern_visualization_apd}, we observe that the attention patterns of these heads are similar to those shown in the main text, with a focus on the temporal dimension. This indicates that these heads are indeed crucial for understanding the temporal dynamics of video content.

\subsection{Statistical Analysis Details}
\label{apd:correlation_analysis_details}

\begin{figure}[t]
    \centering
    \includegraphics[width=0.95\columnwidth]{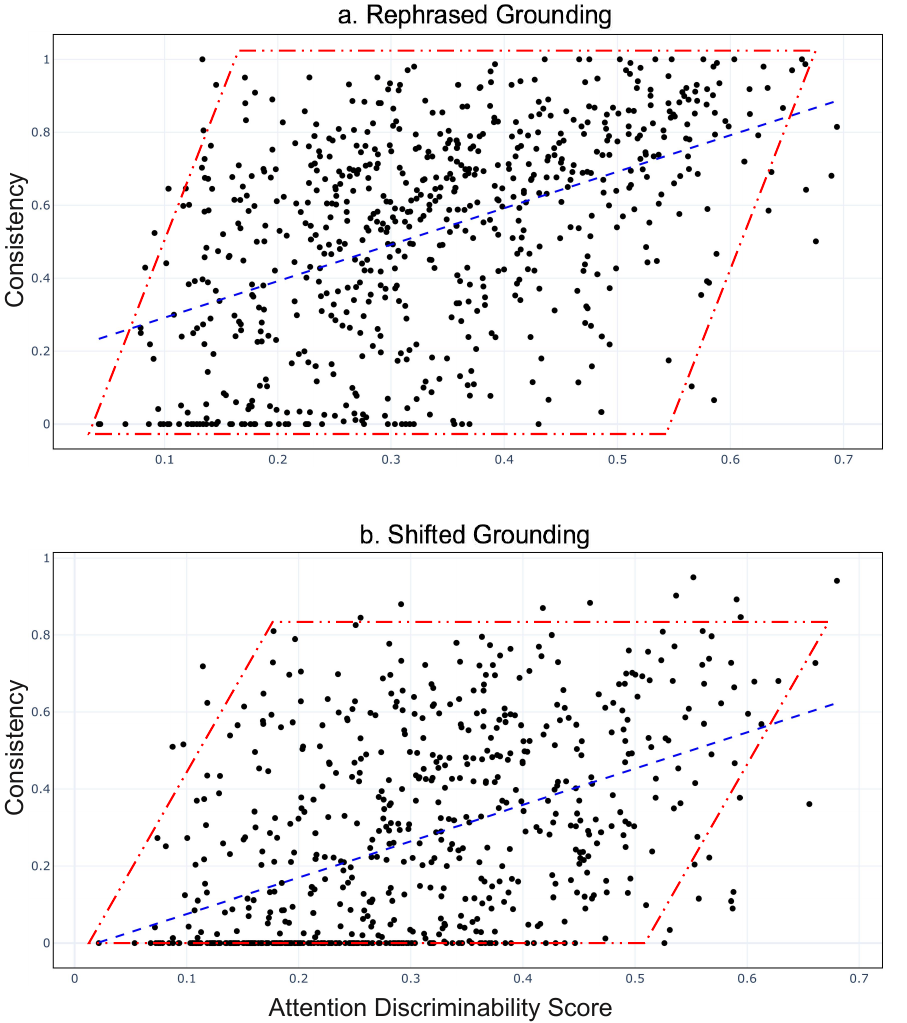}
    \caption{Scatter plot showing the relationship between the consistency score and the attention discriminability score, with a least-squares regression line included.}
    \label{fig:corr_scatter}
\end{figure}

\begin{figure}[t]
    \centering
    \includegraphics[width=0.95\columnwidth]{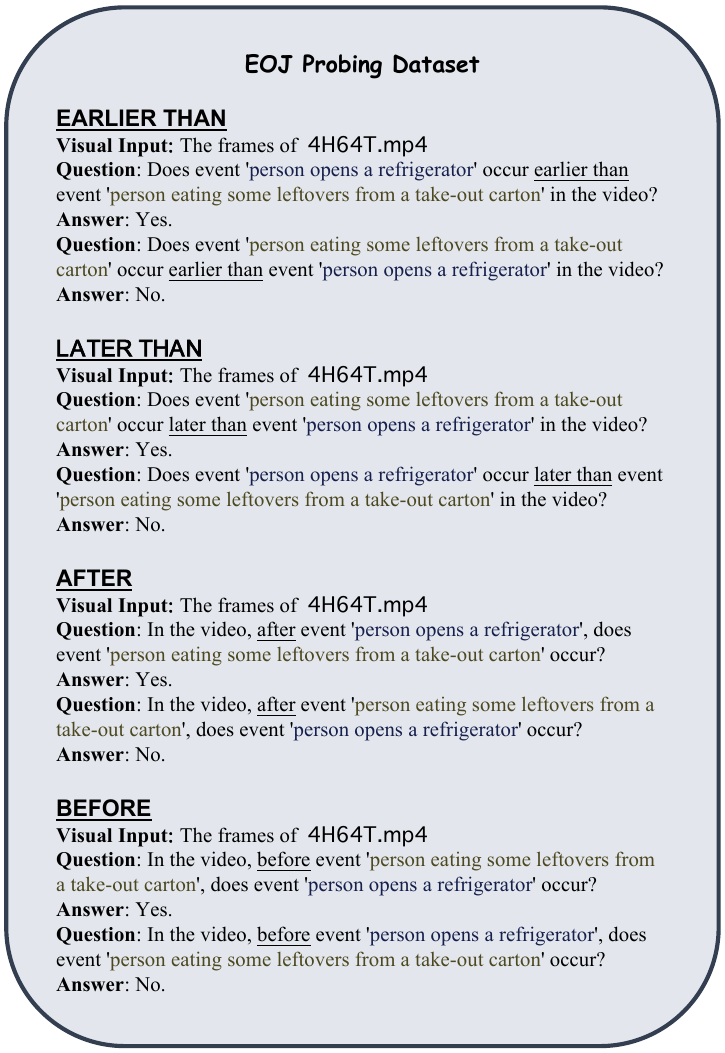}
    \caption{An example of the EOJ Probing dataset. Each sample contains two events with a clear order, and four sets of eight logically equivalent questions are constructed for their order relationship.}
        \label{fig:EOJ_example}
\end{figure}

\begin{figure}
    \centering
    \includegraphics[width=0.95\columnwidth]{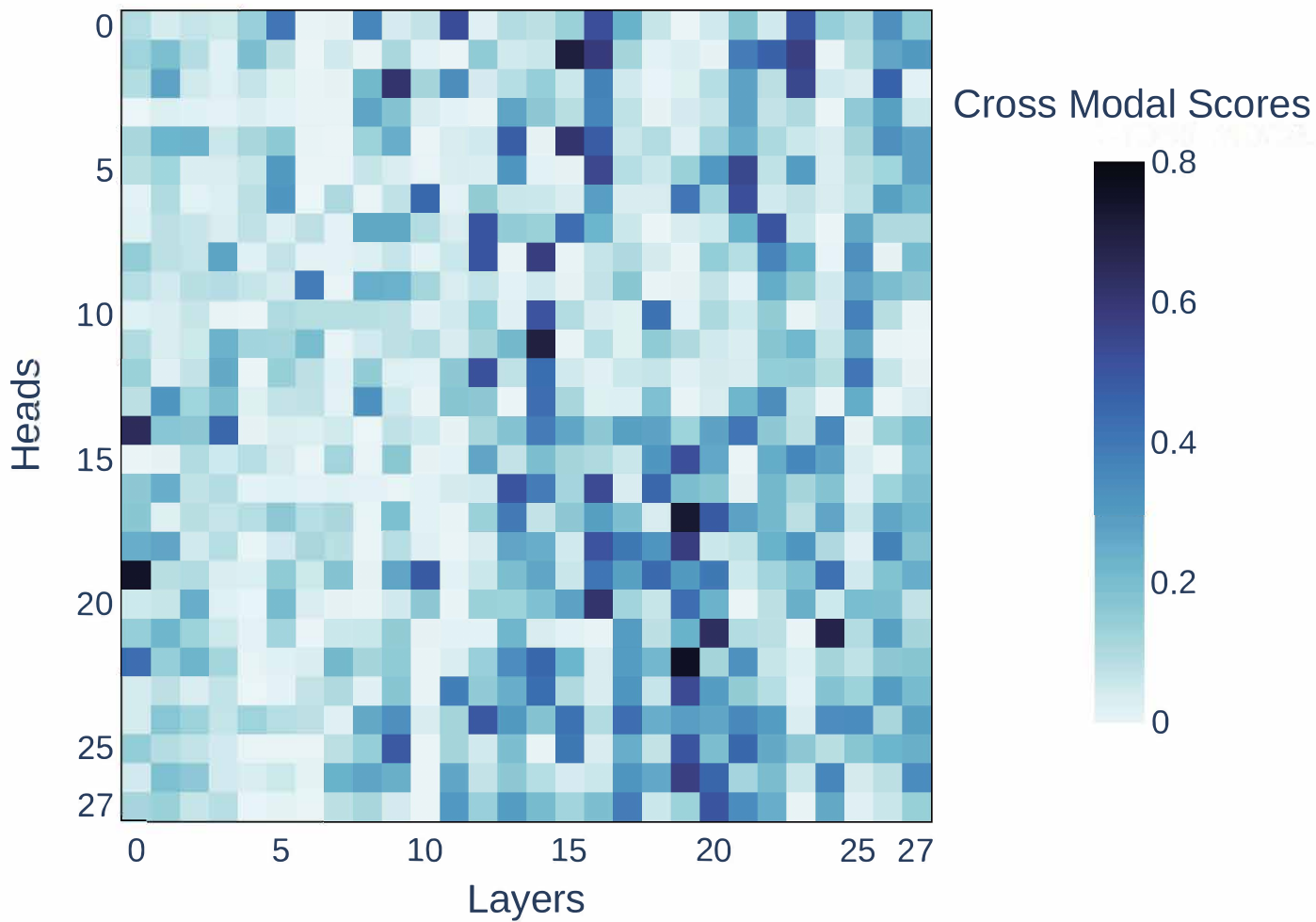}
    \caption{The distribution of cross-modal attention heads in Qwen2.5-VL. The x-axis represents the attention layer index, and the y-axis represents the head index.}
    \label{apd:fig:cross_modal_attention_heads}
\end{figure}
In the main text, we conduct the correlation analysis of attention discriminability scores and consistency scores in TimeChat. To further support the analysis in Section 2.1, we provide the scatter plot with the least-squares regression line to illustrate the relationship between these two metrics.

As shown in Figure~\ref{fig:corr_scatter}, we observe that as the consistency score increases, the distribution of attention discriminability scores shifts to the right. The points clustered on the left side of the x-axis in the figure correspond to difficult samples where the model's predicted IoU is 0. 

For rephrased grounding and shifted grounding, the Pearson correlation coefficients between the two metrics are $0.4778$ and $0.4788$, respectively, with p-values of $\mathbf{5.18 \times 10^{-41}}$ and $\mathbf{3.69 \times 10^{-41}}$, both far below $\mathbf{0.05}$, indicating a statistically significant positive correlation between the two metrics. Since Video-LLMs are inherently complex nonlinear systems, the figure shows that the two metrics are not linearly correlated. However, their positive correlation is statistically significant, as evidenced by the p-values.

\subsection{Details of Analysis on EOJ Task}
\label{apd:generalizability_analysis}

To explore the correlation between attention discriminability and consistency performance on the Event Order Judgement (EOJ) task, we constructed a probing dataset based on the annotations from Charades-STA test subset, containing 1,000 consistent question-answer pairs. Each pair includes a video clip and multiple logically equivalent questions that require the model to determine the order of different events in the video clip. An example of the question-answer pairs is shown in Figure~\ref{fig:EOJ_example}.

\subsubsection{Detection and Interpretation}
Using the same method as in the main text, we first identify the cross-modal attention heads in Qwen2.5-VL that are relevant to the EOJ task. The distribution of these heads is shown in Figure~\ref{apd:fig:cross_modal_attention_heads}. We observe that, consistent with the analysis in the main text, the cross-modal scores in Qwen2.5-VL are also primarily concentrated in a few heads in the middle layers.

Next, we visualize the attention patterns of key heads on several samples, as shown in Figure~5 in main text. We find that the attention patterns of these heads are consistent with the analysis in the main text, establishing a mapping relationship between the query tokens of event queries and the video key tokens corresponding to the time range of events through attention scores, thus achieving cross-modal temporal relationship alignment.

\subsubsection{Correlation Analysis}
In the EOJ task, the consistency score \( c^{v} \) is defined as follows:
\begin{equation}
    c_{v} = \frac{\sum_{q \in Q_{v}} F1(r_{v,q})}{|Q_{v}|}, \quad v \in \mathcal{D}
\label{eq:consistency_score_eoj}
\end{equation}
where $c_{v}$ indicates the model's \textbf{EOJ Consistency Score} for video sample $v$, $r_{v,q}$ denotes the model's response to question $q$ for video sample $v$, $Q_{v}$ refers to the set of questions associated with video sample $v$, and $\mathcal{D}$ represents the dataset of all video samples.

Considering that the EOJ task involves multiple events, the \textbf{EOJ Attention Discriminability Score} is defined as the Kullback-Leibler divergence~\cite{kldiv} between the average attention score distributions of two events as follows:
\begin{gather}
    P^{h,v}_{e} = \frac{\sum_{q \in T_{e}} A^{h,v}_{q,V}}{\sum_{k \in V} \sum_{q \in T_{e}} A^{h,v}_{q,k}} \text{,}
    \label{eq:attention_distribution_eoj} \\
    S^{h,v}_{eoj,disc.} = P^{h,v}_{e_{1}} \log \frac{P^{h,v}_{e_{1}}}{P^{h,v}_{e_{2}}} + P^{h,v}_{e_{2}} \log \frac{P^{h,v}_{e_{2}}}{P^{h,v}_{e_{1}}} \text{,}
    \label{eq:attention_discriminability_eoj} \\
    S^{v}_{eoj,disc.} = \frac{1}{|H_{t}|} \sum_{h \in H_{t}} S^{h,v}_{eoj,disc.} \text{,}
    \label{eq:average_attention_discriminability_eoj}
\end{gather}
where $P^{h,v}_{e}$ denotes the average attention score distribution of all text tokens of event $e$ to the set of all visual tokens $V$, $T_{e}$ represents the set of text tokens of event $e$, $A^{h,v}_{q,V}$ is the attention score from text token as query token $q$ to all visual tokens $V$ as key tokens, $S^{h,v}_{eoj,disc.}$ is the EOJ temporal discriminability score of attention head $h$ for video sample $v$ containing two events $e_{1}$ and $e_{2}$, $H_{t}$ means the set of the top $t$ cross-modal score attention heads, and $S^{v}_{eoj,disc.}$ is the average eoj temporal discriminability score of the top $t$ cross-modal attention heads for video sample $v$.

\begin{figure}[t]
    \centering
    \includegraphics[width=0.9\columnwidth]{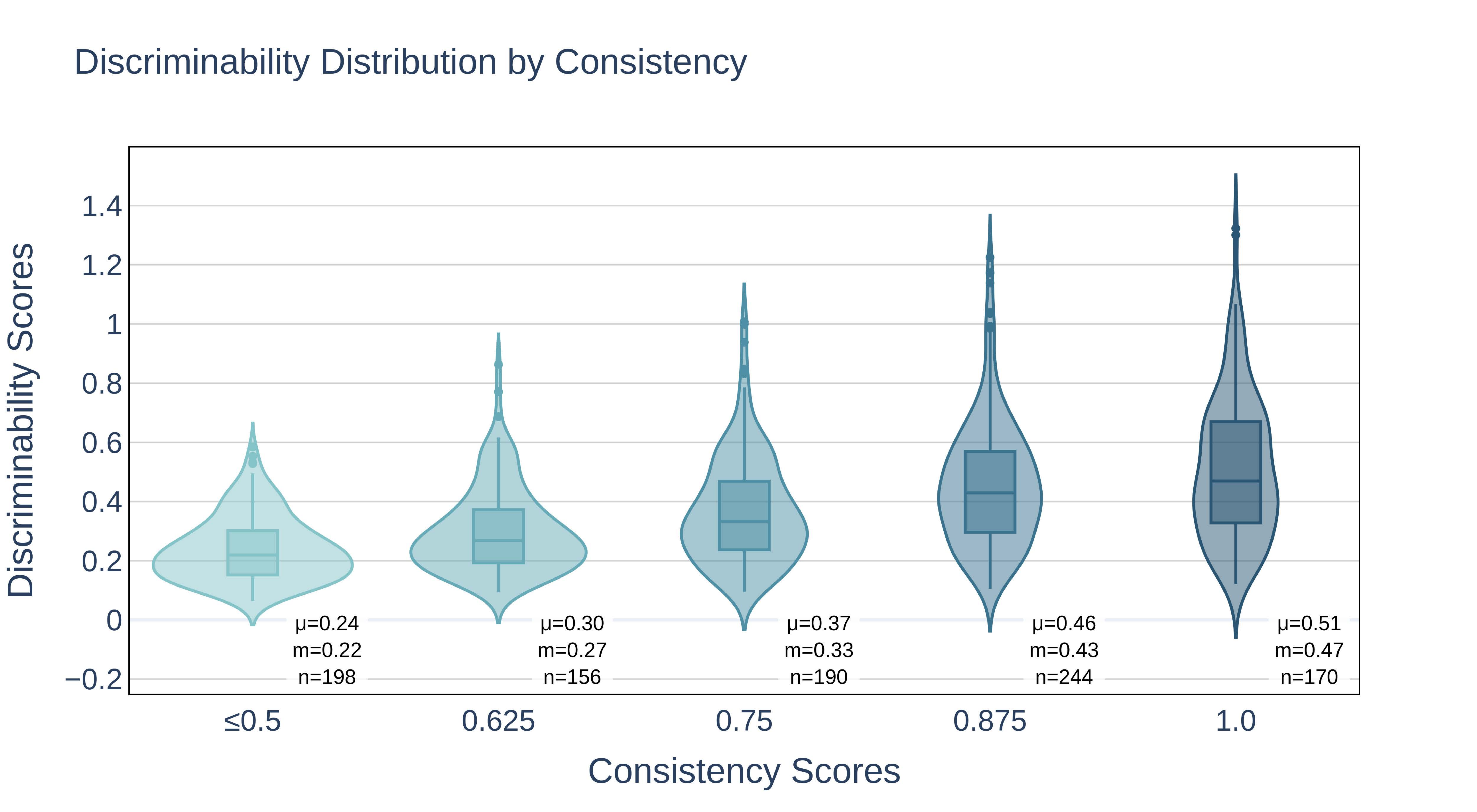}
    \caption{Violin plot of analysis results on EOJ task.}
    \label{fig:qwen_eoj_violin}
\end{figure}

\begin{figure}[t]
    \centering
    \includegraphics[width=0.95\columnwidth]{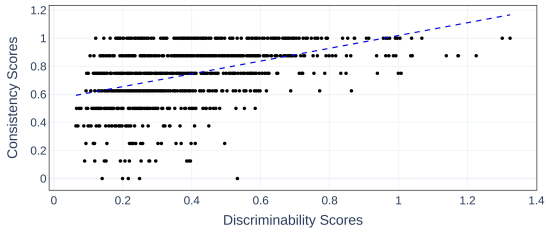}
    \caption{Scatter plot showing the relationship between the consistency score and the attention discriminability score on EOJ task, with a least-squares regression line included.}
    \label{fig:EOJ_correlation_scatter} 
\end{figure}

The correlation analysis of attention discriminability scores and consistency scores is shown in Figure~\ref{fig:EOJ_correlation_scatter} as a scatter plot with the least-squares regression line. We observe that as the consistency score increases, the attention discriminability score also increases significantly, indicating a clear positive correlation between attention discriminability and consistency performance. The corresponding violin plot is shown in Figure~\ref{fig:qwen_eoj_violin}.

In conclusion, our analysis is validated across both the Qwen2.5-VL model and the EOJ task, underscoring the robustness and generalizability of our conclusions. This provides compelling evidence that attention temporal discriminability is a key factor of performance for Video-LLMs in general temporal understanding tasks.

\begin{table}[t]
\centering
\setlength{\tabcolsep}{1.5pt}
\fontsize{9pt}{9pt}\selectfont
{    
    \begin{tabular}{cccccc}
    \toprule
    \multirow{2}{*}{\textbf{Models}}  & \multicolumn{5}{c}{\textbf{Consistency Metrics}}  \\
        & G & R-Ground & S-Ground & H-Verify & C-Verify \\

    \midrule
    \multicolumn{6}{c}{\textbf{Charades-CON}} \\
    \midrule
    GPT-4o 
    &  {28.5} &  {21.2 (74.3)} &  {9.3 (32.8)} &  {17.8 (62.4)} &  {20.3 (71.3)} \\
    gemini-1.5 
    &  {34.6} &  {29.7 (85.7)} &  {24.8 (71.7)} &  {22.8 (65.8)} &  {24.5 (70.8)} \\
    \midrule
     \multicolumn{6}{c}{\textbf{ActivityNet-CON}}  \\
    \midrule
    GPT-4o
    &  {26.8} &  {18.1 (67.5)} &  {10.4 (38.8)} &  {16.5 (61.7)} &  {18.4 (68.8)} \\

    gemini-1.5 
    &  {37.8} &  {30.8 (81.4)} &  {24.8 (65.6)} &  {22.4 (59.3)} &  {26.8 (70.8)} \\

    \bottomrule
    \end{tabular}
}
    \caption{Results of other models without tuning, listed for reference only~\cite{consistency}.}
    \label{tab:other_baselines}
\end{table}

\section{More Details about Experiments}
\label{apd:experiments}

\subsection{Backbone Models Introduction}
\label{apd:backbone_models}

\textbf{Qwen2.5-VL}~\cite{qwenvl} is a representative Video-LLMs featuring a standard cross-modal architecture. We select Qwen2.5-VL to further validate our findings from the previous section. \textbf{Video-LLaMA}~\cite{videollama} is a classic video-LLM using Qformer and video Qformer for video-text alignment. \textbf{TimeChat}~\cite{timechat}, a SOTA video-LLM for the VTG task, integrates timestamp information into its architecture, enhancing its ability to understand the temporal dynamics of video content. We include Video-LLaMA and TimeChat to ensure a fair comparison with the current SOTA method.

\begin{figure}[t] 
    \centering
    \includegraphics[width=0.9\columnwidth]{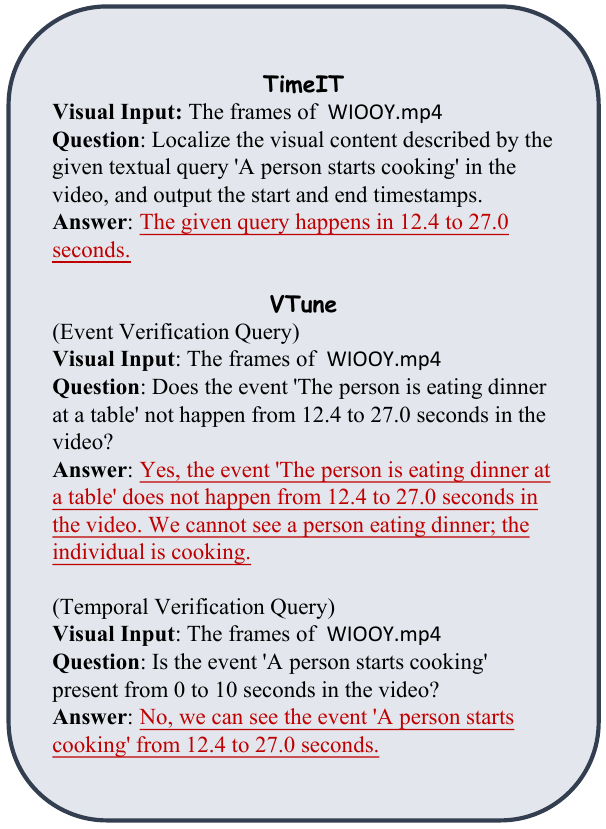}
    \caption{Examples from the training datasets (TimeIT and VTune) used in our training experiments.}
    \label{fig:datasets_example}
\end{figure}
\subsection{Datasets Introduction and Examples}
\label{apd:datasets_example}
\paragraph{TimeIT}\textit{TimeIT} is a dataset proposed by Ren et al.~\cite{timechat} for the various temporal understanding tasks including description, video grounding and temporal reasoning. The splits Charades-VTG and ActivityNet-VTG of it is used in our experiments. These two splits contain 9,848 video clips with 12,408 question-answer pairs and 14,420 video clips with 35,692 question-answer pairs respectively.

\paragraph{VTune}\textit{VTune} is a dataset reported as a method in the original paper~\cite{consistency}. By constructing event verification queries and temporal verification queries annotation data, it effectively enhances the consistency of Video-LLMs from a view of synthetic data augmentation. Two splits Charades-VTune and ActivityNet-VTune contains 9,848 video clips with 99,244 question-answer pairs and 14,420 video clips with 205,510 question-answer pairs.

\paragraph{CON}\textit{Charades-CON} and \textit{ActivityNet-CON} are two benchmarks proposed by Jung et al.~\cite{consistency} to evaluate the consistency of Video-LLMs in temporal understanding tasks. These benchmarks contain 500 video clips with 707 question-answer pairs and 1,422 question-answer pairs. 

The examples of TimeIT and VTune are shown in Figure~\ref{fig:datasets_example}. These two datasets are used for training, while the last two datasets are used for evaluation in our experiments. 

\begin{figure*}[t]
    \centering
    \includegraphics[width=0.9\linewidth,height=0.52\linewidth]{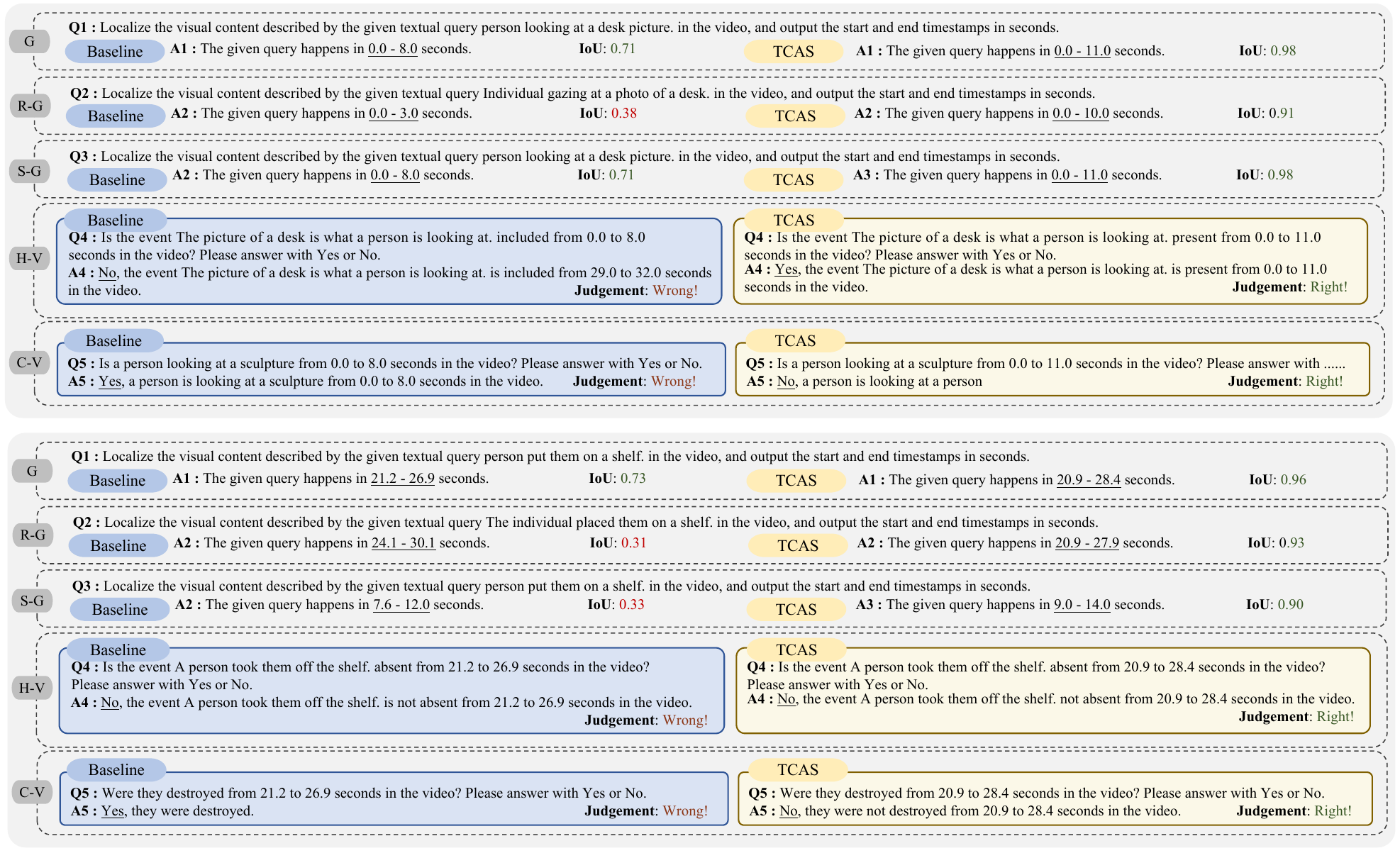}
    \caption{Responses of TimeChat for sample 9T11N (Top) and T0HLX (Bottom) before and after TCAS enhancement.}
    \label{apd:fig:case_study}
\end{figure*}

\subsection{Evaluation Metrics}
\label{apd:evaluation_metrics}
\subsubsection{Evaluation Metrics Details}
\label{apd:evaluation_metrics_details}
Following the previous works~\cite{consistency}, we use these metrics to evaluate the performance of our method:

\begin{itemize}
    \item \textbf{IoU}: The ratio of the intersection over the union between the predicted time range and the ground-truth time range.
    \item \textbf{Accuracy}: The percentage of questions that can be answered correctly by the model. We use it evaluate absolute performance on the verification tasks (H-V. and C-V.) of CON datasets.
    \item \textbf{R@1,IoU=0.5}: The percentage of the IoU of prediction is greater than 0.5. We use it to evaluate the absolute performance of the model on all grounding tasks.
    \item \textbf{R@1,IoU=0.7}: The percentage of the IoU of prediction is greater than 0.7. We use it to evaluate the absolute performance of the model on all grounding tasks.
    \item \textbf{mIoU}: The mean IoU of all questions in the dataset.
    \item \textbf{Consistency}: All Consistency metrics are defined as the original grounding R@1,IoU=0.5 times corresponding absolute performance metrics.
\end{itemize}

\subsubsection{Metrics Rationality}
For analysis in Section~3, we use a sample-wise consistency score (IoU product) for instance-level correlation. In our main experiments (Sec. 5), we follow Jung et al.’s protocol~\cite{consistency}, reporting R@0.5, Acc, and their product with the original grounding score as Appendix Section~\ref{apd:evaluation_metrics_details}, rather than sample-wise score for strictly fair comparison with baselines. For intervention, we report per-setting IoU (Ori/R./S.) to isolate subtask effects. 

The result using sample-wise consistency score (the analysis metric) is shown in Table~\ref{tab:sample_wise_consistency_results}. 
A simple intervention on a small number of attentional heads can slightly improve the consistency score, while low-intensity random intervention leads to a rapid decline in performance.
Gains may look limited because most components are frozen and cannot co-adapt to the intervention, constraining immediate gains. TCAS mitigates this by retraining, allowing the improved attention discriminability to better translate into performance, further confirming the effectiveness of our method TCAS in enhancing temporal consistency.

\begin{table}[t]
\caption{Results of sample-wise consistency score defined as the product of IoU and original grounding score. The $\alpha $ means the intervention intensity. "Random Int." and "Ground Truth Int." mean random intervention and intervention based ground-truth.}
\label{tab:sample_wise_consistency_results}
\centering
  \setlength{\tabcolsep}{1.8pt}
  \fontsize{9pt}{9pt}\selectfont
\begin{tabular}{c|c|ccc|ccccc|c}
\toprule
& \multirow{2}{*}{\makecell[cc]{\textbf{\shortstack{SFT \\ w/o Int.}}}} & \multicolumn{3}{c}{\textbf{Random Int.}} & \multicolumn{5}{c}{\textbf{Ground Truth Int.}} & \multicolumn{1}{c}{\textbf{TCAS}} \\
\cmidrule(lr){3-5} \cmidrule(lr){6-10} \cmidrule(lr){11-11}
 $\alpha$ &  & 0.05 & 0.1 & 0.15 & 0.2 & 0.4 & 0.6 & 0.8 & 1 & -\\
\hline
\rule{0pt}{1.0em}
 $c_{rg.}$  &65.4 & 39.6 & 2.1 & 0 & \underline{66.0} &65.4 &64.1 &62.9 &61.2 & \textbf{70.0}\\
 $c_{sg.}$ &33.1 & 17.2 & 1.8 & 0  & 33.5 &33.7 &34.0 &34.5 & \underline{34.7} & \textbf{37.5} \\
\bottomrule
\end{tabular}

\end{table}

\subsection{Results of Other Baselines and Backbones}
\label{apd:Other_Baselines}
We also compare our method with the following closed-source models: GPT-4o~\cite{gpt4} Gemini-1.5 Flash~\cite{gemini} in table~\ref{tab:other_baselines}. It is worth noting that these results are reported in~\cite{consistency} and are listed here for reference only.

Furthermore, we conducted comparative experiments on the general VTG task using Qwen2.5-VL and Video-LLaMA, with the results presented in Table~\ref{tab:more_comparison_results_VTG}. In contrast to the improvements seen with TimeChat, our method does not consistently enhance performance on the general VTG task for these models. However, the general localization performance remains stable or slightly improves with the enhanced consistency, which we consider acceptable, as the primary motivation of our work is to improve consistency in an interpretable manner.

\begin{table}[t]
\caption{More comparison results for VTG performance on Charades-STA and ActivityNet-Caption.}
\label{tab:more_comparison_results_VTG}
\centering
\setlength{\tabcolsep}{2pt}
\fontsize{9pt}{10pt}\selectfont
      \begin{tabular}{lcccc}
      \toprule
      \multirow{3}{*}{\textbf{Method}} & \multicolumn{2}{c}{\bf Charades-STA} & \multicolumn{2}{c}{\bf ActivityNet-Cap.} \\
      \cmidrule(l{3pt}r{3pt}){2-3} \cmidrule(l{3pt}r{2pt}){4-5}
       & R@1,0.5 & R@1,0.7 & R@1,0.5 & R@1,0.7 \\ \midrule

                    videollama (w. SFT) & 37.1 & 20.1 & 34.3 & 19.1 \\
        \rowcolor{lightgray}
        \bf videollama (w. TCAS) & \underline{37.5} & 20.1 & \underline{35.2} & \underline{20.0} \\ 
        \midrule

                Qwen2.5vl (w. SFT) & 27.3 & 14.1 & 16.5 & 7.8 \\

        \rowcolor{lightgray}
        \bf Qwen2.5vl (w. TCAS) & \underline{32.4} & \underline{16.6} & 16.5 & \underline{7.9} \\ 

      \bottomrule
      \end{tabular}

\end{table}

\subsection{Case Study}
\label{apd:case_study}
In this section, we present more prediction examples to illustrate the effectiveness of our method in enhancing temporal consistency in Video-LLMs. Our model successfully grounds key events and answers questions about the video, demonstrating improved temporal capabilities and consistency compared to baseline models.

\subsection{Further Ablation Studies on Sensitivity}
\label{apd:further_ablation_studies}

\begin{figure}[t]
    \centering
    \includegraphics[width=0.8\columnwidth]{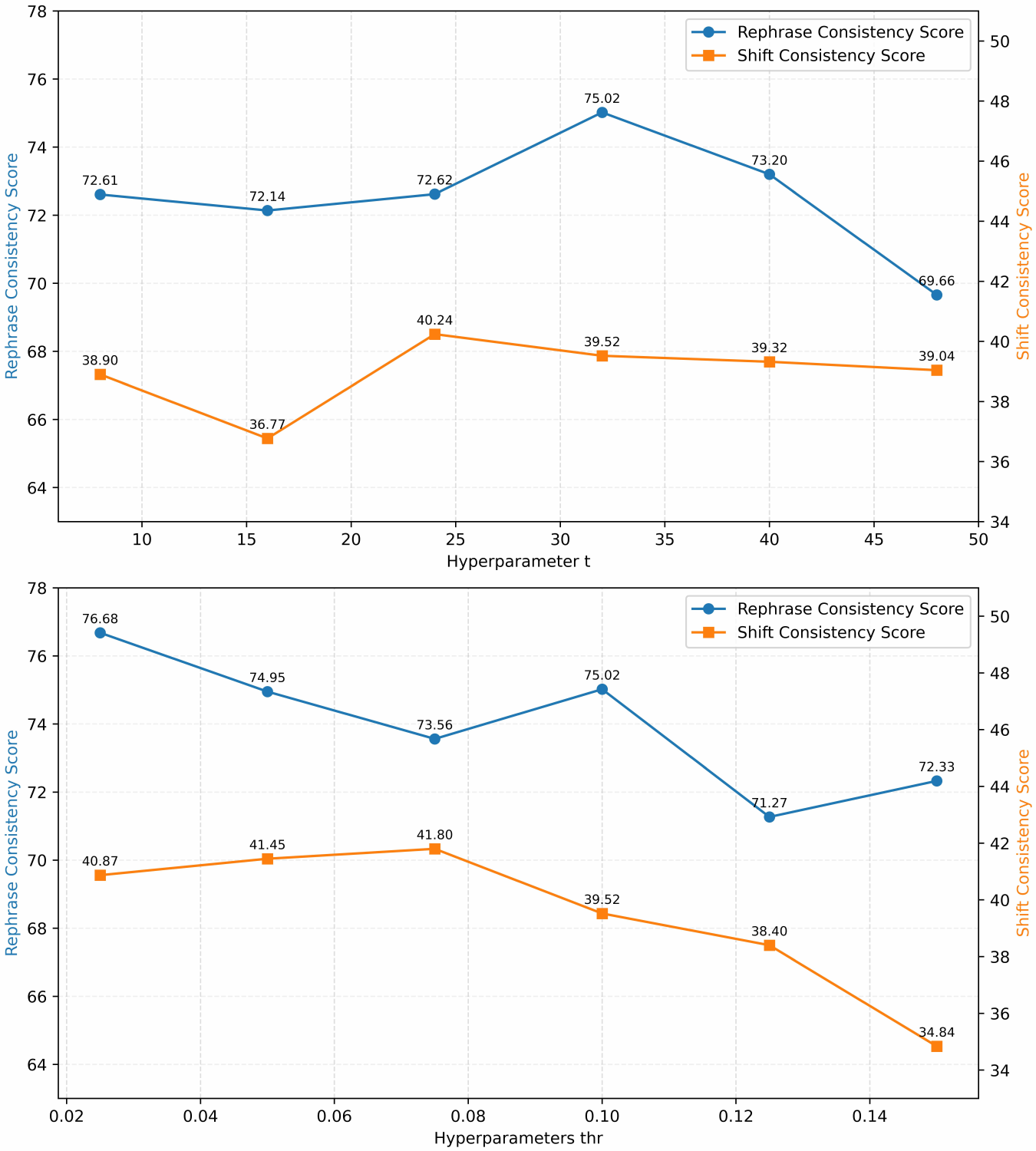}
    \caption{Performance sensitivity analysis of hyperparameters \( t \) (top subplot) and \( thr \) (bottom subplot).}
    \label{fig:sensitivity_apd}
\end{figure}

We conducted a more fine-grained sensitivity analysis for the relatively sensitive hyperparameters \( t \) and \( thr \). The performance variation with respect to these hyperparameters is shown in Figure~\ref{fig:sensitivity_apd}. From the figure, we observe that: 1) As \( t \) increases, both consistency scores exhibit an overall trend of first increasing and then decreasing, emphasizing the need for precise regulation of attention heads, as excessive intervention on heads weakly related to temporal discriminability can have negative effects. 2) When \( thr \) is small, both Rephrase Consistency Score and Shift Consistency Score are relatively high; however, when \( thr \) exceeds 0.1, performance rapidly declines. Because larger values of \( thr \) result in too few tokens being adjusted, which fails to effectively enhance attention discriminability.


\end{document}